\documentclass[10pt,twocolumn,letterpaper]{article}

\usepackage[pagenumbers]{cvpr}   

\usepackage[table]{xcolor}

\usepackage{graphicx}
\usepackage{epsfig}
\usepackage{amsmath}
\usepackage{amssymb}
\usepackage{booktabs}
\usepackage{colortbl}
\usepackage{textcomp}
\usepackage{subcaption}
\usepackage{standalone}
\usepackage{pgfplots} 
\usepackage{multirow}
\usepackage{svg}
\usepackage{siunitx}
\usepackage{lipsum}
\usepackage{pifont}
\usepackage{yfonts}
\usepackage{listings}
\usepackage{algorithm}
\usepackage{algpseudocode}
\usepackage{makecell}
\usepackage[T1]{fontenc}
\usepackage[normalem]{ulem}
\usepackage{nicematrix}
\usepackage{placeins}
\usepackage{circledsteps}
\usepackage{scalerel}

\usetikzlibrary{backgrounds}
\usetikzlibrary{positioning}
\usetikzlibrary{quotes}
\usetikzlibrary{patterns}
\usetikzlibrary{spy}
\usetikzlibrary{arrows,arrows.meta}
\usetikzlibrary{decorations.pathmorphing,decorations.shapes}
\usetikzlibrary{shapes.geometric}
\usetikzlibrary{shapes.symbols}
\usetikzlibrary{patterns}
\usetikzlibrary{calc}

\newcommand{\VEC}[1]{\mathbf{#1}}          
\newcommand{\VECG}[1]{\boldsymbol{#1}}     

\newcommand{\putindex}[3]{\vtop{\hbox{\hspace{#3} $#1$}
            \hbox{\raise 6mm \hbox{$\scriptscriptstyle #2$}}}}

\newcommand{\gradx}[0]{\vtop{\hbox{\rm grad}
            \hbox{\raise 2.5mm \hbox{\rm \hspace{2mm} \footnotesize x}}}}

\newcommand{\grady}[0]{\vtop{\hbox{\rm grad}
            \hbox{\raise 2.5mm \hbox{\rm \hspace{2mm} \footnotesize y}}}}

\newcommand{\grad}[1]{\vtop{\hbox{\rm grad}
            \hbox{\raise 2.5mm \hbox{#1}}}}

\newcommand{\PROB}[0]{{\rm P}}        

\newcommand{\stz}{\rule{0mm}{2.3ex}}

\newcommand{\stzdown}{\rule[-1.2ex]{0mm}{3.5ex}}

\newcommand{\btb}{     \begin{tabbing}             }
\newcommand{\bte}{     \end{tabbing}               }

\definecolor{tu0}{rgb}{0.7451, 0.1176, 0.2353}

\definecolor{tu1}{rgb}{1.0000, 0.8039, 0.0000}
\definecolor{tu11}{rgb}{1.0000, 0.8627, 0.3020}
\definecolor{tu12}{rgb}{1.0000, 0.9020, 0.4980}
\definecolor{tu13}{rgb}{1.0000, 0.9412, 0.6980}
\definecolor{tu14}{rgb}{1.0000, 0.9608, 0.8000}

\definecolor{tu2}{rgb}{0.9804, 0.4314, 0.0000}
\definecolor{tu21}{rgb}{0.9882, 0.6039, 0.3020}
\definecolor{tu22}{rgb}{0.9882, 0.7137, 0.4980}
\definecolor{tu23}{rgb}{0.9922, 0.8275, 0.6980}
\definecolor{tu24}{rgb}{0.9961, 0.8863, 0.8000}

\definecolor{tu3}{rgb}{0.6902, 0.0000, 0.2745}
\definecolor{tu31}{rgb}{0.7529, 0.2000, 0.4196}
\definecolor{tu32}{rgb}{0.8431, 0.4980, 0.6353}
\definecolor{tu33}{rgb}{0.9216, 0.7490, 0.8196}
\definecolor{tu34}{rgb}{0.9529, 0.8510, 0.8902}

\definecolor{tu4}{rgb}{0.4863, 0.8039, 0.9020}
\definecolor{tu41}{rgb}{0.6431, 0.8627, 0.9333}
\definecolor{tu42}{rgb}{0.7412, 0.9020, 0.9490}
\definecolor{tu43}{rgb}{0.8431, 0.9412, 0.9686}
\definecolor{tu44}{rgb}{0.8980, 0.9608, 0.9804}

\definecolor{tu5}{rgb}{0.0000, 0.5020, 0.7059}
\definecolor{tu51}{rgb}{0.3020, 0.6510, 0.7961}
\definecolor{tu52}{rgb}{0.5490, 0.7765, 0.8667}
\definecolor{tu53}{rgb}{0.7490, 0.8745, 0.9255}
\definecolor{tu54}{rgb}{0.8510, 0.9255, 0.9569}

\definecolor{tu6}{rgb}{0.0000, 0.3255, 0.4549}
\definecolor{tu61}{rgb}{0.2510, 0.4941, 0.5922}
\definecolor{tu62}{rgb}{0.5490, 0.6941, 0.7529}
\definecolor{tu63}{rgb}{0.7490, 0.8314, 0.8627}
\definecolor{tu64}{rgb}{0.8510, 0.8980, 0.9176}

\definecolor{tu7}{rgb}{0.0314, 0.0314, 0.0314}
\definecolor{tu71}{rgb}{0.3725, 0.3725, 0.3725}
\definecolor{tu72}{rgb}{0.5882, 0.5882, 0.5882}
\definecolor{tu73}{rgb}{0.7529, 0.7529, 0.7529}
\definecolor{tu74}{rgb}{0.8667, 0.8667, 0.8667}

\definecolor{tu8}{rgb}{0.7765, 0.9333, 0.0000}
\definecolor{tu81}{rgb}{0.8431, 0.9529, 0.3020}
\definecolor{tu82}{rgb}{0.8863, 0.9647, 0.4980}
\definecolor{tu83}{rgb}{0.9333, 0.9804, 0.6980}
\definecolor{tu84}{rgb}{0.9569, 0.9882, 0.8000}

\definecolor{tu9}{rgb}{0.5373, 0.6431, 0.0000}
\definecolor{tu91}{rgb}{0.6784, 0.7490, 0.3020}
\definecolor{tu92}{rgb}{0.7686, 0.8196, 0.4980}
\definecolor{tu93}{rgb}{0.8588, 0.8941, 0.6980}
\definecolor{tu94}{rgb}{0.9059, 0.9294, 0.8000}

\definecolor{tu10}{rgb}{0.0000, 0.4431, 0.3373}
\definecolor{tu101}{rgb}{0.3020, 0.6118, 0.5373}
\definecolor{tu102}{rgb}{0.5490, 0.7490, 0.7020}
\definecolor{tu103}{rgb}{0.7490, 0.8588, 0.8353}
\definecolor{tu104}{rgb}{0.8549, 0.9176, 0.9059}

\definecolor{tu110}{rgb}{0.8000, 0.0000, 0.6000}
\definecolor{tu111}{rgb}{0.8706, 0.3490, 0.7412}
\definecolor{tu112}{rgb}{0.9216, 0.6000, 0.8392}
\definecolor{tu113}{rgb}{0.9608, 0.8000, 0.9216}
\definecolor{tu114}{rgb}{0.9804, 0.8980, 0.9608}

\definecolor{tu120}{rgb}{0.4627, 0.0000, 0.4627}
\definecolor{tu121}{rgb}{0.5961, 0.2510, 0.5961}
\definecolor{tu122}{rgb}{0.7294, 0.4980, 0.7294}
\definecolor{tu123}{rgb}{0.8392, 0.6980, 0.8392}
\definecolor{tu124}{rgb}{0.9216, 0.8510, 0.9216}

\definecolor{tu130}{rgb}{0.4627, 0.0000, 0.3294}
\definecolor{tu131}{rgb}{0.6118, 0.3020, 0.5333}
\definecolor{tu132}{rgb}{0.7569, 0.5490, 0.6980}
\definecolor{tu133}{rgb}{0.8667, 0.7490, 0.8314}
\definecolor{tu134}{rgb}{0.9216, 0.8510, 0.9020}
\DeclareMathOperator*{\argmax}{arg\,max}

\newcommand{\src}[0]{{\mathcal{D}^\mathrm{S}}}
\newcommand{\tgt}[0]{{\mathcal{D}^\mathrm{T}}}

\newcommand{\network}[1]{{\texttt{#1}}}

\newcommand{\basemodel}[1]{\raisebox{.8pt}{\textcircled{\raisebox{-0.9pt}{#1}}}}

\newcommand{\datatrain}[0]{{\mathcal{D}_\mathrm{train}}}
\newcommand{\datafull}[0]{{\mathcal{D}_\mathrm{full}}}

\newcommand{\datateststar}[0]{{\mathcal{D}_\mathrm{test*}}}
\newcommand{\datadev}[0]{{\mathcal{D}_\mathrm{dev}}}

\newcommand{\csteststar}[0]{{\mathcal{D}^\mathrm{CS}_\mathrm{test*}}}
\newcommand{\csdev}[0]{{\mathcal{D}^\mathrm{CS}_\mathrm{dev}}}

\newcommand{\bddteststar}[0]{{\mathcal{D}^\mathrm{BDD}_\mathrm{test*}}}

\newcommand{\mvteststar}[0]{{\mathcal{D}^\mathrm{MV}_\mathrm{test*}}}

\newcommand{\acdcteststar}[0]{{\mathcal{D}^\mathrm{ACDC}_\mathrm{test*}}}

\newcommand{\kittiteststar}[0]{{\mathcal{D}^\mathrm{KIT}_\mathrm{test*}}}

\newcommand{\gtavtrain}[0]{{\mathcal{D}^\mathrm{GTA5}_\mathrm{train}}}

\newcommand{\gtavdev}[0]{{\mathcal{D}^\mathrm{GTA5}_\mathrm{dev}}}

\newcommand{\synthiatrain}[0]{{\mathcal{D}^\mathrm{SYN}_\mathrm{train}}}

\newcommand{\synthiadev}[0]{{\mathcal{D}^\mathrm{SYN}_\mathrm{dev}}}

\pgfplotsset{
table/search path={{./}{../}},
compat = 1.3
}
\graphicspath{{./fig}{../fig}}
\newcommand{\minus}{\scalebox{0.75}[1.0]{$-$}}
\newcommand{\rot}[1]{\rotatebox{90}{#1}}
\newcommand{\na}{\makecell[c]{-}}
\newcolumntype{g}{>{\columncolor{tu73}}c}
\newcolumntype{b}{>{\columncolor{tu14}}c}
\newcolumntype{a}{>{\columncolor{tu74}}c}
\newcolumntype{h}{>{\columncolor{tu74!70!white}}c}

\usepackage[pagebackref=true,breaklinks=true,letterpaper=true,colorlinks,bookmarks=false]{hyperref}
\usepackage[capitalize]{cleveref}
\crefname{section}{Sec.}{Secs.}
\crefname{section}{Section}{Sections}
\crefname{subsection}{Section}{Sections}
\crefname{table}{Table}{Tables}
\crefname{table}{Tab.}{Tabs.}

\def\cvprPaperID{} 
\def\confName{}
\def\confYear{}

\begin{document}
\title{A Re-Parameterized Vision Transformer (ReVT) \\ for Domain-Generalized Semantic Segmentation}
\author{Jan-Aike Termöhlen	\qquad Timo Bartels \qquad Tim Fingscheidt\\
Technische Universitat Braunschweig, Germany\\
{\tt\small \{j.termoehlen, timo.bartels, t.fingscheidt\}@tu-bs.de}
}
\maketitle
\begin{abstract}
 The task of semantic segmentation requires a model to assign semantic labels to each pixel of an image. However, the performance of such models degrades when deployed in an unseen domain with different data distributions compared to the training domain. We present a new augmentation-driven approach to domain generalization for semantic segmentation using a re-parameterized vision transformer (\network{ReVT}) with weight averaging of multiple models after training. We evaluate our approach on several benchmark datasets and achieve state-of-the-art mIoU performance of $47.3\%$ (prior art: $46.3\%$) for small models and of $50.1\%$ (prior art: $47.8\%$) for midsized models on commonly used benchmark datasets. At the same time, our method requires fewer parameters and reaches a higher frame rate than the best prior art. It is also easy to implement and, unlike network ensembles, does not add any computational complexity during inference.\footnote{Code is available at \href{https://github.com/ifnspaml/ReVT}{https://github.com/ifnspaml/ReVT}}
\end{abstract}
\section{Introduction}
\label{sec:intro}
Many methods for machine perception, e.g., for semantic segmentation, employ deep neural networks (DNNs)~\cite{fingscheidt_dnndataautomateddriving}. Due to the high labeling cost for semantic segmentation data, more and more synthetic data are used for training these DNNs. After training on the labeled (source) domain they should operate as robustly as possible in similar, but unseen (target) domains. However, this is often not the case since the data of the target domain differ from those of the training domain, leading to a so-called domain gap. There are many methods to deal with this domain gap that either require samples from the target domain during training~\cite{Schwonberg2023Survey,Bolte2019a,araslanov2021self}, or alter the target data or the network parameters during inference~\cite{Klingner2020c,Klingner2020d,Termoehlen2021}.
An approach that does not have these drawbacks is domain generalization (DG).
The aim of domain generalization is to train a network in a way that it generalizes well to unseen domains without any adaptation steps. 
\begin{figure}[t!]
  \centering
  \includegraphics{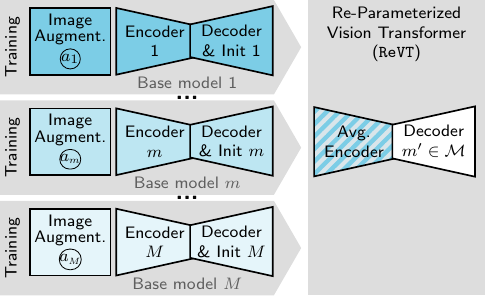}
  \caption{\textbf{High-level overview} of the generalization method. The set of all $M$ base models trained on individual augmentations \raisebox{2pt}{\Circled[inner ysep=5pt,inner xsep=0pt]{$a_{\scaleto{m}{2pt}}$}} is denoted by $\mathcal{M}\!=\!\{1,...,m,...,M\}$. For the re-parameterized vision transformer (\network{ReVT}), any decoder $m' \in \mathcal{M}$ can be used.}
  \label{fig:00-eye-catcher}
  \vspace{-4.5mm}
\end{figure}
Although neural networks that employ vision transformer encoders currently achieve the best performance in segmentation tasks, modern DG methods are mostly presented with \network{ResNet}-based models, such as \network{DeepLabv3+}\cite{Chen2018a} and \network{FCN}~\cite{Long2015}. Due to its strong performance with a comparable or smaller number of parameters, we employ the transformer-based \network{SegFormer}~\cite{Xie2022segformer} as the baseline for our domain generalization method.\par
A training or post-processing method that has proven itself in many applications is re-parameterization. Here, either individual layers, e.g., convolutional layers, or entire models trained with potentially different augmentations can be averaged to improve performance and generalization of the final model. The averaging can be performed either during training \cite{Saemann2022,SWA} or after training \cite{yolov7,Wortsman2022modelsoups}. 
As sketched in \autoref{fig:00-eye-catcher}, in our work we advantageously combine the strengths of selected image augmentations \cite{Hendrycks2022pixmix,Tomasi1998bilateral,Xie2022segformer} with the re-parameterization and show that this method leads to a significantly better generalization capability for transformer-based models. We also show that the method does not improve the performance of the commonly used \network{ResNet}-based models when trained with standard stochastic gradient descent (SGD), but that this can be overcome by the use of the AdamW optimizer~\cite{Loshchilov2019}. \par
As shown in \autoref{fig:00-eye-catcher}, first, $M$ base models are trained, with pre-trained encoders but different random decoder seeding, and potentially with dissimilar augmentations \raisebox{2pt}{\Circled[inner ysep=5pt,inner xsep=0pt]{$a_{\scaleto{m}{2pt}}$}}. Afterwards, the encoder networks can be averaged into one new encoder (re-parameterization), which extracts better generalizing features. This encoder can then be combined with any of the previously trained decoders and be used directly for segmentation. \par
Our contribution with this work is fourfold. First, we propose a re-parameterized vision transformer \network{ReVT} for domain-generalized semantic segmentation, resulting from $M$ augmentation-individual base models. We achieve higher mIoU on unseen domains compared to methods that employ \network{ResNet}-based models, while requiring fewer parameters and achieving higher frame rates than the best prior art. Second, we analyze the effect of different network architectures, network parts, layer types, and optimizers on the re-parameterization. Third, we report on two more real datasets as common in the field and also going beyond customs in the field, we follow a stringent divison of data splits into training, development, and test set. Finally, we set a new state-of-the-art benchmark on the synthetic-to-real domain generalization task for semantic segmentation.
\section{Related Work}
In this section, we discuss related works for our single-source domain generalization method. 
We start with the task of domain generalization, followed by related work on image augmentation and model re-parameterization.
\subsection{Domain Generalization (DG)}
In domain generalization for semantic segmentation, a model is trained on a set of labeled data from a specific (source) domain $\src$ and then evaluated on new data from unseen (target) domains $\tgt$. The goal is to train a model that can generalize well to different domains and accurately segment new images. 
Following Qiao \etal~\cite{Qiao2020}, we distinguish between domain generalization and \textit{single-source} domain generalization. The main difference between these two is that in the former, the model can be exposed to multiple domains during training, e.g., multiple labeled source domains or additional auxiliary domains. A dataset often used as an auxilliary domain is ImageNet~\cite{Deng2009}, which is used to learn the style of real images~\cite{Yue2019,Huang2021}. In the \textit{single-source} domain generalization task, the model is trained solely on one single domain.\par
Muandet \etal~\cite{Muandet2013} proposed a so-called domain-invariant component analysis (DICA) minimizing the dissimilarity across domains during training.
  Li\etal~\cite{Li2017c} learn a domain-agnostic model on multiple domains via low-rank parameterized CNNs.
  Zhang \etal~\cite{Zhang2020a} employ meta-learning for domain generalization and an adaptation of batch norm statistics in the target domain, and therefore present no pure DG method.
  Li \etal~\cite{Li2019c} propose an episodic training with a simple approach of aggregating data from multiple source domains for training.
  Yue \etal~\cite{Yue2019} first randomize the images with the style from real domains and then also enforce pyramid consistency between different styles. Their approach is not single-source domain, but requires an auxiliary domain for the style transfer. Huang~\etal~\cite{Huang2021} follow a similar approach, but proposed to perform the domain randomization in the frequency domain of the images. Pan~\etal~\cite{Pan2018} proposed a new instance-batch normalization (IBN) that is more robust w.r.t. appearance changes such as color shifts or brightness changes. Choi~\etal~\cite{Choi2021} proposed an advanced loss that uses instance selective whitening.  Peng~\etal~\cite{Peng2022semanticaware} proposed a network that includes semantic-aware normalization (SAN) as well as semantic-aware whitening (SAW). WildNet~\cite{Lee2022wildnet} employs feature stylization with styles from an auxiliary domain and enforces semantic consistency between the segmentation masks of stylized and original images and also between the segmentation masks of stylized images and the labels. Other than previous methods, that either perform checkpoint selection\footnote{cf.\ \url{https://github.com/jxhuang0508/FSDR/issues/2\#issuecomment-910089417}}~\cite{Yue2019,Huang2021} or hyperparameter tuning on evaluation data (official validation sets) of the target domains, we follow a stringent approach with distinct development sets for method design and hyperparameter tuning and perform no checkpoint selection (cf.\ \autoref{sec:metrics}). We also evaluate our approach on additional real domains, some of which represent strong domain shifts (cf.\ \autoref{sec:datasets}), and have not been explored by previous approaches. 
\subsection{Image Augmentation}
Image augmentation techniques\cite{Zhang2018mixup,Cubuk2019autoaugment,Yun2019cutmix,Hendrycks2020augmix,Olsson2021classmix,Hendrycks2022pixmix} aim at improving the performance of DNNs by increasing the variability of the training data. They reduce the risk of overfitting, e.g., to synthetic textures~\cite{Kim2020}, and can improve the generalization capability of the model. Some recent augmentation methods mix full images~\cite{Zhang2018mixup}, parts of images~\cite{Yun2019cutmix}, specific class pixels~\cite{Olsson2021classmix}, or combine the previously mentioned augmentation strategies with other image transformations~\cite{Hendrycks2020augmix,Hendrycks2022pixmix}. 
We propose to use a number of $M$ so-called base models with \textit{individual} augmentations \raisebox{2pt}{\Circled[inner ysep=5pt,inner xsep=0pt]{$a_{\scaleto{m}{2pt}}$}} drawn from PixMix~\cite{Hendrycks2022pixmix}, bilateral filtering~\cite{Tomasi1998bilateral}, and the baseline augmentations from the \network{SegFormer} method~\cite{Xie2022segformer}.
\subsection{Model Re-Parameterization}
\begin{figure}[t!]
  \centering
\includegraphics[width=\columnwidth]{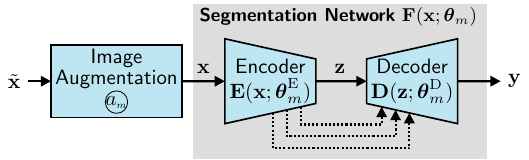}
  \caption{\textbf{Training setup} of base model $m$ and notations. Dotted lines indicate skip connections.}
  \label{fig:00_blockdiagramm_overview}
  \vspace{-4.5mm}
\end{figure}
The stochastic weight averaging (SWA)~\cite{SWA} method averages the network weights of the model during the training process with stochastic gradient descent (SGD) using a cyclical or constant learning rate. Similarly, Sämann~\etal~\cite{Saemann2022} also employ the model averaging during training. A related method was also investigated by Kamp~\etal~\cite{Kamp2019} as an efficient decentralized learning protocol.\par
In contrast to methods that employ the re-parameterization during the training process~\cite{SWA,Tarvainen2017meanteacher,Saemann2022}, we adopt the re-parameterization approach by Wortsman \etal~\cite{Wortsman2022modelsoups} that the authors dubbed ``model soups'' and performed the averaging of the model weights after various training processes.
Note that it is also possible to re-parameterize specific layers and alter the architecture after re-parameterization, e.g., with RepVGG~\cite{Ding2021_repvgg}. Wang~\etal~\cite{yolov7} analyzed these re-parameterization strategies for convolutional layers in different networks and proposed an advanced planned re-parameterized model. 
\section{Proposed Method}
In this section, we will describe the mathematical notations and our new re-parameterized vision transformer (\network{ReVT}) training, including augmentations.
\subsection{Mathematical Notations}
A high-level overview of the employed training setup is given in \autoref{fig:00_blockdiagramm_overview}. During training in the labeled source domain $\src$, an image $\tilde{\VEC{x}}$ is subject to augmentation methods and then denoted as $\mathbf{x} \in \mathbb{G}^{H \times W \times C}$, where $\mathbb{G}$ denotes the set of integer gray values, $H$ and $W$ the image height and width in pixels, and $C\!=\!3$ the number of color channels. 
The augmented images $\VEC{x}$ are then transformed by the segmentation network $\VEC{F}$ with network parameters $\VECG{\theta}$ to obtain an output tensor $\mathbf{y} = \VEC{F}({\VEC{x}};\VECG{\theta}) = (y_{i,s})\in \mathbb{I}^{H \times W \times S}$ that contains a pixel-wise posterior probability $y_{i,s} = \PROB(s|i,{\VEC{x}})$ for all classes $s \in \mathcal{S}$ at each pixel index $i \in \mathcal{I} = \{1,2,...,H\cdot W\}$, with $\mathbb{I}=[0,1]$. 
The segmentation network consists of an encoder $\VEC{z}=\VEC{E}(\VEC{x};\VECG{\theta}^\mathrm{E})$ and a decoder (segmentation head) $\VEC{y}=\VEC{D}(\VEC{z};\VECG{\theta}^\mathrm{D})$, with the parameters $\VECG{\theta}^\mathrm{E}$ and $\VECG{\theta}^\mathrm{D}$, respectively, resulting in $\VEC{y}=\VEC{F}(\VEC{x};\VECG{\theta})=\VEC{D}(\VEC{E}(\VEC{x};\VECG{\theta}^\mathrm{E});\VECG{\theta}^\mathrm{D}) $. The number of parameters in a parameter tensor is denoted as $|\VECG{\theta}|$. Different parameter tensors $\VECG{\theta}_m$ for the same architecture are marked by a subscript $m\in\mathcal{M}$, where $\mathcal{M}=\{1,2,...,M\}$ is the respective index set and $M$ is the total number of models. 
The set of classes $\mathcal{S} = \{1,2,...,S\}$ contains the same $S$ classes for source domain training and target domain inference (closed set). To obtain the final classification map $\mathbf{m}=(m_{i})\in \mathcal{S}^{H \times W}$, we compute ${m}_i = \argmax_{s\in\mathcal{S}} {y}_{i,s}$. 
\subsection{Re-Parameterized Vision Transformer (\textbf{\network{ReVT}})}
\begin{figure}[t!]
\centering
\includegraphics{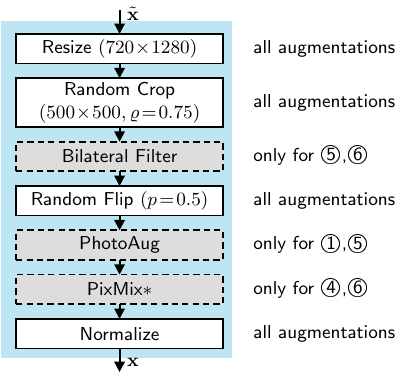}
  \caption{\textbf{Image augmentation pipeline} from \textcolor{red}{Figures}~\ref{fig:00-eye-catcher} and \ref{fig:00_blockdiagramm_overview} employed during training. Use of blocks for augmentations \basemodel{$a$} noted at the side (cf. \autoref{network_arch}).}
  \label{fig:00-augmenation-pipeline}
  \vspace{-4.5mm}
\end{figure}
To the best of our knowledge we are the first to introduce vision transformer re-parameterization to domain generalization for semantic segmentation. In particular, each of the base models has seen an individual augmentation strategy in training. An illustrated overview of our proposed single-source domain generalization method for semantic segmentation is given in \autoref{fig:00-eye-catcher}. First, $M$ segmentation networks of the same architecture are trained using ImageNet-pretrained encoders and different random decoder seeds and potentially also different augmentation strategies.\par
\textbf{Image augmentation}:
The base model-individual augmentation steps employed during training are an important component of our method. We have illustrated the image augmentation pipeline in \autoref{fig:00-augmenation-pipeline}. The baseline augmentation pipeline consists of resizing, random cropping, random flipping (Random Flip), photometric augmentation (PhotoAug), followed by a normalization to zero mean and unit variance. 
The bilateral filter~\cite{Tomasi1998bilateral} can be inserted before the random flipping (\autoref{fig:00-augmenation-pipeline}, upper gray box).
While PhotoAug is our default, it can optionally be replaced by the PixMix~\cite{Hendrycks2022pixmix} algorithm (\autoref{fig:00-augmenation-pipeline}, lower gray box). The original PixMix algorithm applies randomly selected augmentations. We employ the baseline augmentations (Random Flip + PhotoAug) here, which is why we refer to our PixMix variant as PixMix*. All non-self-explanatory augmentations are explained in more detail in Supplement \autoref{supp:augmenations}. As a result of either different random seeding or augmentation, the network parameters will differ after training (\autoref{fig:00-eye-catcher}, left side). \par
\textbf{Re-parameterized vision transformer (\textbf{\network{ReVT}})}:
After the training, the model weights $\VECG{\theta}_m$, $m\in\mathcal{M}$, can be averaged resulting in $ \VECG{\theta}_\diamond$.
The new averaged model weights $\VECG{\theta}_\diamond$ could be used during inference. 
Different to the method described by Sämann \etal~\cite{Saemann2022}, \textit{we only re-parameterize the encoder weights} 
\begin{equation}
  \label{eq:1}
  \VECG{\theta}_\diamond^\mathrm{E} = \frac{1}{M}\sum_{m\in\mathcal{M}}\VECG{\theta}_m^\mathrm{E} \, ,
\end{equation} 
resulting in our proposed \network{ReVT} as
\begin{equation}
  \label{eq:2}
  \VEC{y}^{\network{ReVT}}=(y_{i,s}^{\network{ReVT}})=\VEC{D}(\VEC{E}^{\network{ReVT}}(\VEC{x};\VECG{\theta}_\diamond^\mathrm{E});\VECG{\theta}_{m'}^\mathrm{D}) \, ,
\end{equation}
with an arbitrarily chosen decoder $m'\in\mathcal{M}$.
\section{Experimental Setup}
In the following, we introduce the employed
datasets and network architectures. Afterwards, we explain the training and evaluation settings, as well as the evaluation metrics. All architectures, procedures, and metrics are
implemented using \texttt{PyTorch}~\cite{Paszke2019} and the \texttt{MMSegmentation} toolbox~\cite{mmseg2020}.
\subsection{Datasets}
\label{sec:datasets}
 \begin{table}[t!]
  \centering
  \renewcommand{\arraystretch}{1.}
  \setlength{\tabcolsep}{.65em}
  \caption{\textbf{Employed datasets}. The synthetic datasets GTA5~\cite{Richter2016} and SYNTHIA~\cite{Ros2016} are used as (single) source domains ($\src$). We employ various real-world datasets as target domains ($\tgt$) to show the generalization capability of the proposed method.}
  
    \setlength{\tabcolsep}{.25em}
    \begin{tabular}{@{}lrrrr@{}}\toprule
       \multirow{1}{*}{\makecell[l]{\textbf{Dataset}\\\textbf{Name}}} & \multicolumn{4}{c}{\textbf{\# Images in}} \\
         & $\datafull$ & $\datatrain$ & $\datadev$ & $\datateststar$ \\
        \midrule
        GTA5~\cite{Richter2016}&  24,966 & 12,403 & 6,382 & \na\\
       
        SYNTHIA~\cite{Ros2016} (SYN) & 9,400 &  6,580  & 2,820 & \na\\
        \midrule
       
        Cityscapes~\cite{Cordts2016} (CS)& \na &  \na  & 500 & 500 \\
        Mapillary Vistas~\cite{Neuhold2017} (MV) & \na &  \na  & \na & 2,000 \\
        BDD100k~\cite{Yu2018b} (BDD) & \na &  \na  & \na & 1,000 \\
        ACDC~\cite{Sakaridis2021acdc} & \na &  \na  & \na & 406 \\
        KITTI~\cite{AbuAlhaija2018} (KIT) & \na &  \na  & \na & 200 \\ 
        \bottomrule     
    \end{tabular}

  \label{tab:datasets} 
   \vspace{-4.5mm}
\end{table}
In our experiments we evaluate multiple established domain generalization benchmarks for semantic segmentation. The definition of the individual splits and their respective number of images is shown in \autoref{tab:datasets}. As our synthetic domains we employ GTA5~\cite{Richter2016} and SYNTHIA~\cite{Ros2016}. We employ the three commonly used real-world datasets Cityscapes~\cite{Cordts2016}, BDD100k~\cite{Yu2018b}, and Mapillary Vistas~\cite{Neuhold2017} as target domains. Different to other publications, we also employ the ACDC~\cite{Sakaridis2021acdc} and the KITTI~\cite{AbuAlhaija2018} datasets to provide more evidence of domain generalization on real domains. Particularly the ACDC dataset offers considerable benefit, since it includes images from four adverse conditions (fog, nighttime, rain, and snow), which are not present in the synthetic data. 
In DG benchmarks, there is no common practice on choosing which part of the synthetic dataset to use for training. Some publications use the entire GTA5 or SYNTHIA dataset for training~\cite{Yue2019,Huang2021}. Other publications use the official training split of GTA5 and define their own training split for SYNTHIA~\cite{Choi2021,Lee2022wildnet}. We follow Choi \etal~\cite{Choi2021} and employ their training and development split for SYNTHIA, and the official GTA5 training and validation set for training and development, respectively. Most DG methods base their design decisions on the official validation sets of the target domains and do not report test results. Since this approach is not rigorous, we follow an approach from domain adaptation~\cite{araslanov2021self} and sample 500 random images from the (unused) Cityscapes training set to be used as our development set, see \autoref{tab:datasets}. To allow comparison, we use the official validation sets of the real domains as test sets. To avoid confusion with the official (partly unpublished) test sets, we name our test sets ``test$*$''.
\subsection{Network Architectures}
\label{network_arch}
\begin{table}[t!]
  \centering
  \renewcommand{\arraystretch}{1.}
  \setlength{\tabcolsep}{.45em}
  \caption{Models and corresponding \textbf{number of parameters} for the full \textbf{segmentation networks} employed in this paper.}

    \begin{tabular}{@{}llc@{}}
        \toprule
        \makecell[l]{\textbf{Segmentation}\\\textbf{Network}}&\makecell[l]{\textbf{Encoder}}& \makecell{$|\VECG{\theta}|$ \\\textbf{($\cdot 10^6$)} }\\
        \midrule
        \multirow{2}{*}{\network{DeepLabv3+}~\cite{Chen2018a}}&\texttt{ResNet50} &$43.7$\\
        &\texttt{ResNet101} &$62.7$\\
        \midrule
        \multirow{3}{*}{\network{SegFormer}~\cite{Xie2022segformer}} &\texttt{MiT-B2} &$27.4$\\
        &\texttt{MiT-B3} &$47.2$\\
         &\texttt{MiT-B5} &$84.7$\\
        \bottomrule
    \end{tabular}

  \label{tab:number-of-parameters-full-networks} 
   \vspace{-4.5mm}
\end{table}
For our experiments we employ two different network architectures that use an encoder-decoder structure, as illustrated in \autoref{fig:00_blockdiagramm_overview}. The employed segmentation networks and the corresponding number of parameters are listed in \autoref{tab:number-of-parameters-full-networks}. First, we use a \network{SegFormer}~\cite{Xie2022segformer} architecture with multiple skip connections from early layers to the decoder (\network{SegFormer} head). Second, a \network{DeepLabv3+}~\cite{Chen2018a} with only one skip connection from an early layer to the decoder is investigated. To ensure comparability with other reference methods, we will also perform experiments with several encoder sizes.
If only \network{SegFormer} is mentioned and no additional information is given, this shall refer to the use of an \network{MiT-B5} encoder. If only \network{DeepLabv3+} is mentioned and no additional information is given, this shall refer to the use of a \network{ResNet-101} encoder. For the re-parameterization of the models several encoders are required. As can be seen in \autoref{fig:00-eye-catcher}, the $M$ models that are used in this process will be referred to as \textit{base models} (not to be confused with baseline models, which are simply the standard \network{SegFormer} or \network{DeepLabv3+} models), each with a potentially different image augmentation. The different image augmentations \raisebox{2pt}{\Circled[inner ysep=4.5pt,inner xsep=0pt]{$a_{\scaleto{m}{2pt}}$}} for each base model $m$ are identified by \basemodel{1}, \basemodel{2}, etc. If the same image augmentations is used multiple times, \eg, $\raisebox{2pt}{\Circled[inner ysep=3.5pt,inner xsep=0pt]{$a_{\scaleto{1}{4pt}}$}}\!=\!\raisebox{2pt}{\Circled[inner ysep=3.5pt,inner xsep=0pt]{$a_{\scaleto{2}{4pt}}$}}\!=\!\raisebox{2pt}{\Circled[inner ysep=3.5pt,inner xsep=0pt]{$a_{\scaleto{3}{4pt}}$}}\!=\!\basemodel{1}$, then the $M\!=\!3$ base models were just trained with a different random seed. We then denote the used augmentations by \{\basemodel{1},\basemodel{1},\basemodel{1}\}.

\subsection{Training, Evaluation, and Metrics}
\label{sec:metrics}
The hyperparameters for the image augmentation, training and evaluation (inference) procedures are provided in Supplement \autoref{supp:settings}. \par
Unlike other methods~\cite{Yue2019,Huang2021}, we do not use the test$*$ sets (official validation sets) of the individual target domains for hyperparameter tuning or selection of training checkpoints. We train all our models for a fixed number of iterations and evaluate the checkpoint from the last iteration. 
We want to emphasize that we firmly believe that this is closer to a realistic deployment if a domain generalization method. To evaluate the methods, we employ the standard mean intersection over union (mIoU) of 19 segmentation classes~\cite{Cordts2016,Richter2016,Sakaridis2021acdc}. Hyperparameter tuning is only based on our (self-defined) development sets $\datadev$, see \autoref{tab:datasets}. Specifically, we employ the mIoU mean on our out-of-domain (OOD) development sets for our design decisions on the proposed \network{ReVT}. 
\begin{table}[t]
  \centering
  \renewcommand{\arraystretch}{0.9}
  \setlength{\tabcolsep}{.65em}
  \caption{Performance (mIoU (\%)), when \textbf{different network parts} are used in the \textbf{re-parameterization}. \textbf{Training} was performed on the \textbf{GTA5} ($\src\!=\!\gtavtrain$) training set. \textbf{Evaluation} is performed on the \textbf{Cityscapes development set} ($\tgt\!=\!\csdev$). Reported is the mean mIoU of \{\basemodel{1},\basemodel{1},\basemodel{1}\} models. For the re-parameterization, the mean is computed with one averaged encoder and the three associated decoders $m\in\{1,2,3\}$. Best results in bold face, second-best underlined.}
  
    \begin{tabular}{@{}clc@{}}
        \toprule
        \makecell{\textbf{Segmentation}\\\textbf{Network}}  
     &\makecell[l]{\textbf{Method:}\\\textbf{Re-Parameterization ...}}  &\makecell{\textbf{mIoU (\%)}\\\textbf{on $\csdev$}}  \\
        \midrule
       \multirow{4}{*}{\makecell{\network{SegFormer}\\\small(\network{MiT-B5})}} & ... not done (Baseline) & $\underline{44.3}$ \\
        & ... in encoder only & $\mathbf{47.5}$\\ 
         & ... in decoder only & $31.5 $\\ 
          & ... in full network & $34.2$\\ 
           \midrule
         \multirow{4}{*}{\makecell{\network{DeepLabv3+}\\(\small\network{ResNet-101})}} & ... not done (Baseline)  & $\mathbf{34.7}$ \\
        & ... in encoder only & $\underline{31.9} $ \\
         & ... in decoder only &  $\hphantom{0}1.9$ \\
          & ... in full network &  $\hphantom{0}1.9$\\
    
        \bottomrule
    \end{tabular}

  \label{tab:reparamlocation} 
  \vspace{-4.5mm}
\end{table}
To compare our method to other reference methods, we also report an mIoU over multiple domains. We follow Lee \etal~\cite{Lee2022wildnet} and Choi \etal~\cite{Choi2021} and evaluate the benchmark (BM) mean mIoU over the following benchmark set of data splits: $\{\gtavdev, \synthiadev, \csteststar, \bddteststar, \mvteststar\}$. 
We report the model size $|\VECG{\theta}|$ and the frame rate in frames per second (fps), as measured on an \network{NVIDIA A100} GPU.\\
\section{Evaluation and Discussion}
\label{sec:experiments}
In this section, we will first investigate the basics of re-parameterization w.r.t.\ re-parameterized network parts, layers, and the number of base models. Afterwards, we evaluate different base model augmentations and optimizer methods during training to design our final \network{ReVT}. Finally, we compare our models to prior art DG methods. \\
\subsection{Basic Investigations on Re-Parameterization}
For the following experiments on basics of network re-parameterization, we only employ base models that were trained with the baseline image augmentation \basemodel{1}. If not stated otherwise, the experiments are performed with $M\!=\!3$ base models (\{\basemodel{1},\basemodel{1},\basemodel{1}\}). Reported is always the mIoU on the Cityscapes development set ($\tgt\!=\!\csdev$). 
\par \textbf{Re-parameterized network parts}:
\begin{table}[t]
  \centering
  \renewcommand{\arraystretch}{0.85}
  \setlength{\tabcolsep}{.15em}
  \caption{Performance (mIoU (\%)) for different optimizer {set\-ups}, \ie, optimizer, learning rate, weight decay, etc. We investigate the effect of the standard \network{SegFormer} optimizer setup (gray rows) and \network{Deeplabv3+} optimizer setup (yellow rows) as shown in the Supplement \autoref{supp:settings}, \autoref{tab:hyperparameters}. \textbf{Training} was performed on the full synthetic \textbf{GTA5} ($\src\!=\!\gtavtrain$) dataset. Evaluation is performed on the \textbf{Cityscapes development set} ($\tgt\!=\!\csdev$). Reported is the mean mIoU of \{\basemodel{1},\basemodel{1},\basemodel{1}\} models. Best results in bold face.}
  
    \begin{NiceTabular}{@{}cccc@{}}[colortbl-like]
        \toprule
       {\textbf{\makecell{Network}}} &  {\textbf{\makecell{Optimizer  \\setup \\ following ...}}}& \multicolumn{2}{c}{\makecell{\textbf{mIoU (\%)} \textbf{on} $\csdev$}}\\ 
      \cmidrule{3-4}
       &\footnotesize(cf.~\autoref{tab:hyperparameters})& Baseline & re-parameterized\\
       \midrule
       \multirow{2}{*}{\network{SegFormer}}&\rowcolor{tu74}\network{SegFormer}&44.3&\textbf{47.5}\\
       \stz&\rowcolor{tu14}\network{DeepLabv3+}&\textbf{46.2}&46.9 \\
       \midrule
       \multirow{2}{*}{\network{DeepLabv3+}}&\rowcolor{tu74}\network{SegFormer}&\textbf{35.3}&\textbf{38.5} \\
       \stz&\rowcolor{tu14}\network{DeepLabv3+}&34.7&31.9 \\
   
       \bottomrule
    \end{NiceTabular}

  \label{tab:optimizer} 
   \vspace{-4.5mm}
\end{table}
In \autoref{tab:reparamlocation} we investigate the effect of the re-parameterization, when applied to different network parts. We compare baseline models (no re-parameterization) and re-parameterization of the encoder only, the decoder only, and the full network. We show results for the \network{SegFormer} as well as for \network{DeepLabv3+}. It can be seen, that the encoder-only re-parameterization is the only setup which improves $3.2\%$ absolute (abs.) over the baseline results from $44.3\%$ to $47.5\%$.
We also see that the \network{DeepLabv3+} does not profit at all from any form of re-parameterization, actually, the performance even degrades from $34.7\%$ to $31.9\%$. Therefore, in the remainder of the paper, we will use the re-parameterization for the vision transformer \network{SegFormer} to obtain the \network{ReVT}. We will also refer to the encoder-only re-parameterization simply as re-parameterization. In the following experiment we will further investigate why the \network{DeepLabv3+} did not profit from re-parameterization and how this effect can be avoided.
\par \textbf{Optimizer choice}:
In \autoref{tab:optimizer} we investigate the performance differences of baseline models and re-parameterized models when trained with different optimizer setups. The optimizer setup comprises all settings regarding the training process. We give a detailed list in Supplement \autoref{supp:settings} in \autoref{tab:hyperparameters}. We test the effect of the standard optimizer setup for the \network{SegFormer} (AdamW, gray rows) and \network{DeepLabv3+} (SGD, yellow rows). It can be seen that the \network{SegFormer} baseline is stronger when trained with the \network{DeepLabv3+} setup ($46.2\%$ vs.\ $44.3\%$), but the gain from re-parameterization becomes significantly smaller ($0.7\%$ abs.\ improvement vs.\ $3.2\%$ abs.\ improvement). For the \network{DeepLabv3+}, the \network{SegFormer} optimizer setup is the much better choice, because on the one hand the baseline has a better performance ($35.3\%$ vs.\ $34.7\%$), and on the other hand, it shows significant improvement ($3.2\%$ abs.) instead of deterioration ($-2.8\%$ abs.). For more analysis, see Supplement~\autoref{supp:add-optimizer-exp}. 
\par \textbf{Re-parameterized blocks/layer types}:
\begin{table}[t]
  \centering
  \renewcommand{\arraystretch}{0.9}
  \setlength{\tabcolsep}{.65em}
  \caption{Performance (mIoU (\%)) of the \network{SegFormer}, when certain enocoder \textbf{block or layer types} are used in the \textbf{re-parameterization}. \textbf{Training} was performed on the full synthetic \textbf{GTA5} ($\src\!=\!\gtavtrain$) dataset. \textbf{Evaluation} is performed on the \textbf{Cityscapes development set} ($\tgt\!=\!\csdev$). Reported is the mean mIoU $\pm$ the standard deviation of \{\basemodel{1},\basemodel{1},\basemodel{1}\} models. For the re-parameterization, the mean $\pm$ standard deviation is computed with one averaged encoder and the three associated decoders $m\in\{1,2,3\}$. Best results in bold face, second-best underlined.}

    \begin{tabular}{@{}lc@{}}
        \toprule
        \makecell[l]{\textbf{Method:}\\\textbf{Re-Parameterization ...}}   & \makecell{\textbf{mIoU (\%)}\\\textbf{on $\csdev$}}  \\
        \midrule
         ... not done (Baseline \network{SegFormer}) & $44.3 \pm 1.9$ \\
         ... in all blocks/layers &$\mathbf{47.5} \pm 0.1$ \\
         ... in patch embedding blocks only & $44.7 \pm 1.7$ \\
         ... in attention blocks only & $45.4 \pm 1.1$ \\
         ... in Mix-FFN blocks only & $\underline{47.0} \pm 0.6$ \\
        \midrule
         ... in convolutional layers only & $45.1 \pm 1.5$ \\
         ... in fully connected layers only & $46.8 \pm 0.6$ \\
        
        \bottomrule
    \end{tabular}

  \label{tab:reparamlayertypes} 
   \vspace{-4.5mm}
\end{table}
In \autoref{tab:reparamlayertypes} we investigate the performance of re-parameterization of different block and layer types within the \network{SegFormer} encoder. For each row, only the stated blocks or layers are re-parameterized, the rest of the models is kept the same for all $m\in\mathcal{M}$. The location of the specific blocks and layers is depicted in Supplement \autoref{supp:block-diagramms}. It can be seen in \autoref{tab:reparamlayertypes} that the method works best when all parameters of the encoder are used in the re-parameterization. The selection of specific blocks or layers does not bring any advantage. However, all independently evaluated layer / block types yield an improvement over the baseline. 
\par
\begin{figure}[t]
  \centering
  \includegraphics{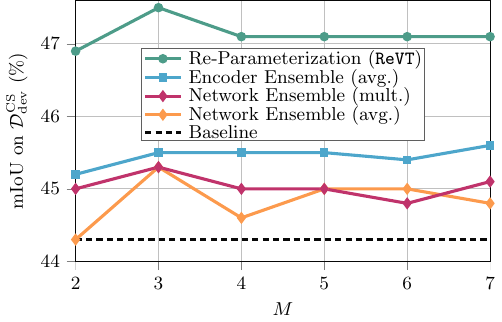}
  \caption{Performance (mIoU (\%)) of the \textbf{re-parameterization} vs.\ \textbf{network/encoder ensembles} for different numbers $M$ of base models. The \textbf{training} of the base models (\network{SegFormer}) was performed on the \textbf{GTA5} ($\src\!=\!\gtavtrain$) dataset. The evaluation is performed on the \textbf{Cityscapes development set} ($\tgt\!=\!\csdev$). The baseline mean is calculated from eight different models \basemodel{1}, and the re-parameterization from $M$ models \basemodel{1}.}
  \label{fig:numofmodels} 
   \vspace{-4.5mm}
\end{figure}

\textbf{Number of base models}:
In \autoref{fig:numofmodels} we show the performance of the re-parameterization vs.\ various ensembling techniques for a different number $M$ of models. For the \network{ReVT} (green), the mIoU is computed with an averaged encoder and all associated $M$ decoders. For the encoder ensemble (blue), the feature maps $\VEC{z}$ from the encoders are averaged and then processed by all associated $M$ decoders. For the network ensemble, the $M$ output posteriors are averaged (orange) or multiplied (red). 
For $M>2$, all methods consistently outperform the baseline (dashed line). In our case, three base models ($M\!=\!3$) provide the best results for the re-parameterization as well as for the network ensemble. The encoder ensemble on the other hand profits from a larger number of base models and yields the best performance for $M\!=\!7$. The re-parameterization outperforms all ensembling techniques for all values of $M$ by at least $1.5\%$ abs. and for $M\!=\!3$ by at least $2.0\%$ abs. It also comes with an $M$-fold lower computational complexity in inference. 
\subsection{\textbf{\network{ReVT}} Method Design}
\begin{table*}[t]
  \centering
  \renewcommand{\arraystretch}{0.9}
  \setlength{\tabcolsep}{.22em}
  \caption{Performance (mIoU (\%)) of the \network{SegFormer} model (with an \network{MiT-B5} encoder) using different domain generalization methods. \textbf{Training} was performed on the synthetic \textbf{GTA5} ($\src\!=\!\gtavtrain$) dataset. \textbf{Evaluation} is performed on the \textbf{Cityscapes}, \textbf{GTA5}, and \textbf{SYNTHIA development sets} (gray columns) and on the \textbf{test$*$ data} of various real-world target datasets ($\tgt\!=\!\datateststar$). 
  Reported is the mean mIoU $\pm$ the standard deviation of $M\!=\!3$ models with various image augmentations. For the \network{ReVT}, the mean $\pm$ standard deviation is computed with one averaged encoder and the three associated decoders $m\in\{1,2,3\}$. Best results in bold face, second-best underlined.
  }

\extrarowheight=\aboverulesep
    \addtolength{\extrarowheight}{\belowrulesep}
    \aboverulesep=0pt
    \belowrulesep=0pt
    \resizebox{\textwidth}{!}{%
    \begin{tabular}{@{}clhgggcccc@{}}
        \toprule
        \stzdown \hphantom{W} & \makecell[l]{\textbf{Method} \\ performed:} &    \multicolumn{8}{c}{\textbf{mIoU (\%) on}}  \\
      \cmidrule{3-10}
        &&$\gtavdev$& $\synthiadev$ & $\csdev$ & \raisebox{-3.2pt}{\shortstack{\textbf{OOD} \\\textbf{mean}}} & $\csteststar$ & $\bddteststar$ &$\mvteststar$  &\raisebox{-3.0pt}{\shortstack{\textbf{test$*$} \\\textbf{mean}}}  \rule[-1.1ex]{0mm}{3.55ex}  \\
       \midrule
       \parbox[t]{2mm}{\multirow{6}{*}{\rotatebox[origin=c]{90}{... during training}}} &Baseline \basemodel{1}& $68.2 \!\pm\! 0.0$ & $33.8 \!\pm\! 0.7$ & $44.3 \!\pm\! 1.9$ & $39.1 \!\pm\! 5.4$ & $45.3 \!\pm\! 1.9$ & $43.3 \!\pm\! 1.3$ & $46.8 \!\pm\! 0.9$ & $45.2 \!\pm\! 2.0$ \\

       &  \minus PhotoAug \basemodel{2}& $\underline{68.5}\!\pm\! 0.1$ & $32.3 \!\pm\! 0.4$ & $42.0 \!\pm\! 1.1$ & $37.2 \!\pm\! 4.9$ & $42.5 \!\pm\! 1.5$ & $42.3 \!\pm\! 0.7$ & $45.5 \!\pm\! 0.9$ & $43.4 \!\pm\! 1.8$ \\
    
       &  \minus PhotoAug, \minus Rand. Flip \basemodel{3}&$\mathbf{69.0}\!\pm\! 0.3$ & $33.0 \!\pm\! 0.5$ & $42.8 \!\pm\! 0.8$ & $37.9 \!\pm\! 4.9$ & $42.3 \!\pm\! 1.3$ & $41.1 \!\pm\! 0.8$ & $46.4 \!\pm\! 0.9$ & $43.3 \!\pm\! 2.5$ \\
       
       & +PixMix*~\cite{Hendrycks2022pixmix} \basemodel{4}&$65.1\!\pm\! 0.2$ & $\mathbf{35.4}\!\pm\! 1.1$ & $\mathbf{46.5}\!\pm\! 0.3$ &$\mathbf{41.0}\!\pm\!5.6$& $\underline{46.9}\!\pm\! 0.8$ & $\underline{46.1}\!\pm\! 1.0$ & $\underline{51.2}\!\pm\! 0.3$ & $\underline{48.1}\!\pm\!2.4$ \\
       
       & +Bilateral Filter (BF)~\cite{Tomasi1998bilateral}\basemodel{5} &$68.0\!\pm\! 0.1$ & $34.3\!\pm\! 0.6$ & $45.7\!\pm\! 0.3$ &$40.0\!\pm\!5.7$& $46.8\!\pm\! 0.5$ & $44.2\!\pm\! 1.1$ & $49.4\!\pm\! 1.0$ & $46.8\!\pm\!2.3$ \\
       
       & +PixMix*~\cite{Hendrycks2022pixmix} +BF~\cite{Tomasi1998bilateral} \basemodel{6}&$64.3\!\pm\! 0.1$ & $\underline{35.2}\!\pm\! 0.3$ & $\underline{46.2}\!\pm\! 0.7$ &$\underline{40.7}\!\pm\!5.5$& $\mathbf{47.5}\!\pm\! 0.8$ & $\mathbf{46.7}\!\pm\! 0.1$ & $\mathbf{51.5}\!\pm\! 0.4$ & $\mathbf{48.6}\!\pm\!2.2$ \\    
     
       \midrule

      \parbox[t]{2mm}{\multirow{8}{*}{\rotatebox[origin=c]{90}{...after training}}} & \network{ReVT} \{\basemodel{1},\basemodel{1},\basemodel{1}\}&$68.6\!\pm\! 0.2$ & $35.5 \!\pm\! 0.4$ & $47.5 \!\pm\! 0.1$ & $41.5 \!\pm\! 6.0$ & $49.3 \!\pm\! 0.1$ & $45.3 \!\pm\! 0.5$ & $49.3 \!\pm\! 0.2$ & $48.0 \!\pm\! 1.9$ \\
       & \network{ReVT} \{\basemodel{2},\basemodel{2},\basemodel{2}\}&$\underline{69.1} \!\pm\! 0.1$ & $34.1 \!\pm\! 0.3$ & $44.4 \!\pm\! 0.5$ & $39.3 \!\pm\! 5.2$ & $44.9 \!\pm\! 0.5$ & $44.1 \!\pm\! 0.4$ & $47.6 \!\pm\! 0.4$ & $45.5 \!\pm\! 1.6$ \\
        & \network{ReVT} \{\basemodel{3},\basemodel{3},\basemodel{3}\}&$\mathbf{69.7}\!\pm\! 0.1$ & $35.0\!\pm\! 0.4$ & $46.0\!\pm\! 0.3$ &$40.5\!\pm\!5.5$& $45.7\!\pm\! 0.2$ & $43.5\!\pm\! 0.3$ & $48.8\!\pm\! 0.2$ & $46.0\!\pm\!2.2$ \\
       & \network{ReVT} \{\basemodel{4},\basemodel{4},\basemodel{4}\}&$65.7\!\pm\! 0.1$ & $36.2\!\pm\! 0.0$ & $\mathbf{48.6}\!\pm\! 0.3$ &$\underline{42.4}\!\pm\!6.2$& $48.8\!\pm\! 0.3$ & $47.5\!\pm\! 0.2$ & $\underline{53.2}\!\pm\! 0.3$ & $49.8\!\pm\!2.5$ \\
       & \network{ReVT} \{\basemodel{5},\basemodel{5},\basemodel{5}\}&$68.5 \!\pm\! 0.1$ & $35.9 \!\pm\! 0.0$ & $47.3 \!\pm\! 0.2$ & $41.6 \!\pm\! 5.7$ & $48.6 \!\pm\! 0.3$ & $45.9 \!\pm\! 0.2$ & $51.2 \!\pm\! 0.1$ & $48.6 \!\pm\! 2.2$ \\
       & \network{ReVT} \{\basemodel{6},\basemodel{6},\basemodel{6}\}&$64.9\!\pm\! 0.0$ & $36.1\!\pm\! 0.2$ & $48.0\!\pm\! 0.2$ &$42.1\!\pm\!6.0$ &$\underline{49.7}\!\pm\! 0.2$ & $\mathbf{48.5}\!\pm\! 0.4$ & $\mathbf{53.5}\!\pm\! 0.1$ & $\mathbf{50.5}\!\pm\!2.1$ \\
        & \network{ReVT} \{\basemodel{4},\basemodel{5},\basemodel{6}\}&$66.4\!\pm\!0.7$ & $\underline{36.9}\!\pm\!0.2$ & $47.9\!\pm\!0.5$ & $\underline{42.4}\!\pm\!5.5$ & $49.5\!\pm\!0.4$ & $\underline{48.1}\!\pm\!0.2$ & $53.1\!\pm\!0.2$ & $\underline{50.2}\!\pm\!2.1$ \\
        & \network{ReVT} \{\basemodel{1},\basemodel{4},\basemodel{6}\}&$66.4\!\pm\! 0.7$ & $\mathbf{37.3}\!\pm\! 0.9$ & $\mathbf{48.6}\!\pm\! 0.8$ &$\mathbf{42.9}\!\pm\!5.7$& $\mathbf{50.0}\!\pm\! 0.5$ & $48.0\!\pm\! 0.3$ & $52.8\!\pm\! 0.2$ & $\underline{50.2}\!\pm\!2.0$ \\

        \bottomrule
    \end{tabular}}

  \label{tab:method-design}
  \vspace{-4.5mm}
\end{table*}
In \autoref{tab:method-design} we evaluate various augmentation methods \basemodel{\raisebox{1pt}{$a$}} (see also \autoref{fig:00-augmenation-pipeline} and Supplement \autoref{supp:augmenations}) to identify strong base models. In the lower part of the table we report some ($M\!=\!3$) combinations $\{\raisebox{2pt}{\Circled[inner ysep=3.5pt,inner xsep=0pt]{$a_{\scaleto{1}{4pt}}$}},\raisebox{2pt}{\Circled[inner ysep=3.5pt,inner xsep=0pt]{$a_{\scaleto{2}{4pt}}$}},\raisebox{2pt}{\Circled[inner ysep=3.5pt,inner xsep=0pt]{$a_{\scaleto{3}{4pt}}$}}\}$ of these base models by our re-parameterization. The gray columns indicate our development sets ($\datadev$), where the light grey column is $\gtavdev$. Since we train on $\gtavtrain$, we select our models on the (dark gray) OOD mean mIoU of $\synthiadev$ and $\csdev$. 
We select those base models for further evaluation in the \network{ReVT} that performed best, or second-, or third-ranked on the out-of-domain $\synthiadev$ and $\csdev$ development sets (OOD mean). \\
It can be seen that base models \basemodel{4}, \basemodel{6}, and \basemodel{5} yield the best-, second-, third-ranked performance ($41.0\%$, $40.7\%$, and $40.0\%$) on $\csdev$ as well as $\synthiadev$ (out-of-domain data), whereas the base models \basemodel{3} and \basemodel{2} yield the best and second-ranked performance ($69.0\%$ and $68.5\%$) on $\gtavdev$ (in-domain data). This is to be expected, as the image augmentation makes it harder to learn in the source domain, but forces the base models to generalize slightly, as can be seen in the improved OOD performance of the base models \basemodel{4}, \basemodel{5}, and \basemodel{6}. In the following, we will report the performance for three different \network{ReVTs}. First, the \network{ReVT} \{\basemodel{1},\basemodel{4},\basemodel{6}\} combines the baseline base model with the two best performing augmentation methods, which leads to the best mean OOD performance ($42.9\%$). Second, the \network{ReVT} \{\basemodel{4},\basemodel{5},\basemodel{6}\} combines the best-, second-, and third-ranked augmentation methods. Third, the \network{ReVT} \{\basemodel{4},\basemodel{4},\basemodel{4}\} combines three base models with the single best augmentation method. The later two achieve the second-ranked performance on the OOD data ($42.4\%$). \par
In the lower part of \autoref{tab:method-design} it can be seen that the best test$*$ performance can be achieved with a \network{ReVT} \{\basemodel{6},\basemodel{6},\basemodel{6}\} leading to a test$*$ mean performance of $50.5\%$. Our best dev set \network{ReVT} \{\basemodel{1},\basemodel{4},\basemodel{6}\} achieves $50.2\%$ as test$*$ mean.
\subsection{Comparison to Prior Art DG Methods}
In \autoref{tab:sota} we compare our method (\network{ReVT}) to prior art methods for domain generalization. We sort methods with respect to their encoder model (Enc.) and give the number of parameters of the full network in the third column. We also indicate whether the methods are trained with only one source domain, or if real auxiliary domains are employed and also report the inference frame rate. Methods are grouped to emphasize that w.r.t.\ the number of parameters $|\theta|$ and w.r.t.\ the frame rate, the \network{MiT-B2}-based \network{ReVT} is competitive to \network{ResNet-50}-based methods (group 1), the \network{MiT-B3}-based \network{ReVT} is competitive to \network{ResNet-101}-based methods (group 2), while the \network{MiT-B5}-based \network{ReVT} builds an own group.
Not all methods report the mIoU values for all datasets. In addition to the commonly used datasets, we also evaluate on ACDC~\cite{Sakaridis2021acdc} and KITTI~\cite{AbuAlhaija2018} to provide more evidence of domain generalization on real domains. \\
\begin{table*}[t] 
  \centering
  \renewcommand{\arraystretch}{0.8}
  \setlength{\tabcolsep}{.3em}
  \caption{Performance (mIoU (\%)) of various domain generalization methods employing different segmentation networks, sorted into three performance groups. \textbf{Training} was performed on the synthetic \textbf{GTA5} ($\src\!=\!\gtavtrain$) dataset. The results marked with $^\circ$ are cited from~\cite{Lee2022wildnet} and with $^*$ are cited from the respective paper. All results without any identifier are simulated. \textbf{Evaluation} is performed on the \textbf{SYNTHIA} and \textbf{GTA5 development sets} and on the \textbf{test$*$ data} of various real-world target datasets ($\tgt\!=\!\datateststar$). BM means benchmark. For our simulations we report mean values over three runs with different seeding. Best performance per group in bold face, second best underlined.}
  
\extrarowheight=\aboverulesep
    \addtolength{\extrarowheight}{\belowrulesep}
    \aboverulesep=0pt
    \belowrulesep=0pt
    \begin{tabular}{@{}cclccccccccccc@{}}
        \toprule[.9pt]
        &\multirow{1}{*}{\rot{\makecell[c]{\textbf{Enc.\hphantom{0}}}}} & \textbf{Method} &  \multirow{1}{*}{\makecell{$|\VECG{\theta}|$\\\textbf{($\cdot 10^6$)}} }&\multirow{1}{*}{\textbf{\makecell{ Single \\[-2pt] Source}}}&\multirow{1}{*}{\textbf{\makecell{ Frame \\[-2pt] Rate \\[-2pt] [fps]}}}&\multicolumn{8}{c}{\textbf{mIoU (\%) on}} \\
        \cmidrule{7-14}
       && & &  && $\csteststar$ & $\bddteststar$ & $\mvteststar$ & $\synthiadev$ & $\gtavdev$ &$\acdcteststar$&$\kittiteststar$&\raisebox{-4.0pt}{\shortstack{\textbf{BM} \\\textbf{mean}}} \rule[-1.2ex]{0mm}{3.65ex} \\
        \midrule
        \parbox[t]{2mm}{\multirow{10}{*}{\rotatebox[origin=c]{90}{\small{{\textbf{Group 1}}}}}} &\parbox[t]{2mm}{\multirow{6}{*}{\rotatebox[origin=c]{90}{\small{\network{ResNet-50}}}}} &Baseline$^\circ$ & 43.7&\ding{51}& 7.9 & 35.16 & 29.71 & 31.29 & 27.97&\underline{71.17}&-&-&39.06\\

        &&IBN-Net$^\circ$~\cite{Pan2018}  & 43.6&\ding{51}&8.4& 36.52 & 34.18 & 38.74 & 30.41&70.78&-&-&42.12\\
        &&RobustNet$^\circ$~\cite{Choi2021}& 43.6  &\ding{51}&8.5& 38.78 & 35.64 & 40.38 & 28.97&70.16&-&-&42.78\\
        &&DRPC*~\cite{Yue2019}& 49.6 &\ding{55}&8.3& 37.42& 32.14& 34.12&- &-&-&-&-\\
        &&SAN+SAW*~\cite{Peng2022semanticaware}& 25.6 &\ding{51}&8.1& 39.75 & 37.34 & 41.86 & 30.79&-&-&-&- \\
        &&WildNet$^\circ$~\cite{Lee2022wildnet}  &  43.6 &\ding{55}&7.9& {44.62} & {38.42} &{46.09} & {31.34}&$\mathbf{71.20}$&-&-&{46.33}\\
        \cmidrule{2-14}
         &{\multirow{4}{*}{\rotatebox[origin=c]{90}{\small{\network{MiT-B2}}}}} &Baseline & 27.4 &\ding{51}&\textbf{12.0}&$41.73$ & $38.77$ & $44.15$ & $31.20$ & ${65.95}$ & $30.20$ & $44.34$ & $44.36$ \\

        &&Ours: \network{ReVT} \{\raisebox{.8pt}{\textcircled{\raisebox{-0.9pt}{4}}},\raisebox{.8pt}{\textcircled{\raisebox{-0.9pt}{4}}},\raisebox{.8pt}{\textcircled{\raisebox{-0.9pt}{4}}}\}   & 27.4 &\ding{51}&\textbf{12.0}& $45.06$ & $40.44$ & $49.46$ & $\mathbf{33.29}$ & $62.57$ & $\mathbf{36.62}$ & $\underline{48.94}$ & $46.16$ \\
        &&Ours: \network{ReVT} \{\basemodel{4},\basemodel{5},\basemodel{6}\}   & 27.4 &\ding{51}&\textbf{12.0}&$\underline{45.55}$ & $\mathbf{43.43}$& $\mathbf{49.91}$ & $\underline{33.16}$ & $63.58$ & $36.66$ & $49.27$ & $\underline{47.13}$ \\

       &&Ours: \network{ReVT} \{\raisebox{.8pt}{\textcircled{\raisebox{-0.9pt}{1}}},\raisebox{.8pt}{\textcircled{\raisebox{-0.9pt}{4}}},\raisebox{.8pt}{\textcircled{\raisebox{-0.9pt}{6}}}\}   & 27.4 &\ding{51}&\textbf{12.0}&$\mathbf{46.27}$ & $\underline{43.29}$ & $\underline{49.84}$ & $\mathbf{33.29}$ & ${63.74}$ & $\underline{36.01}$ & $\mathbf{50.13}$ & $\mathbf{47.29}$ \\
        \midrule[.9pt]
        \parbox[t]{2mm}{\multirow{11}{*}{\rotatebox[origin=c]{90}{\small{{\textbf{Group 2}}}}}} &\parbox[t]{2mm}{\multirow{7}{*}{\rotatebox[origin=c]{90}{\small{\network{ResNet-101}}}}} &Baseline$^\circ$ & 62.7&\ding{51}& 5.1 & 35.73 & 34.06 & 33.42 & 29.06&\underline{71.79}&-&-&40.81\\

        &&IBN-Net$^\circ$~\cite{Pan2018}  & 62.6 &\ding{51}&6.0& 37.68 & 36.64 & 36.75 & 30.84 &70.39&-&-&42.46\\
        &&RobustNet$^\circ$~\cite{Choi2021}& 62.6 &\ding{51}& 6.0& 37.26 & 38.66 & 38.09 & 30.17&70.53&-&-&42.94\\
       & &DRPC*~\cite{Yue2019}& 68.6 &\ding{55}&5.3&42.53& 38.72& 38.05&-&-&-&-&-\\
        &&FSDR*~\cite{Huang2021}& 68.6 &\ding{55}&5.3&44.80 &41.20&43.40&-&-&-&-&- \\
       & &SAN+SAW*~\cite{Peng2022semanticaware}& 44.6&\ding{51}& 5.3&45.33 &41.18& 40.77 &31.84 &-&-&-&-\\
        &&WildNet$^\circ$~\cite{Lee2022wildnet}  & 62.6&\ding{55}&5.1 &{45.79} & {41.73} & {47.08} & {32.51}&\textbf{71.91}&-&-&{47.81}\\
     
        \cmidrule{2-14}
    
       &\parbox[t]{2mm}{\multirow{4}{*}{\rotatebox[origin=c]{90}{\small{\network{MiT-B3}}}}} &Baseline & 47.2&\ding{51}& \textbf{10.7}&$43.92$ & $42.96$ & $46.36$ & $32.57$ & $67.59$ & $34.44$ & $45.18$ & $46.68$ \\
        
       & &Ours: \network{ReVT} \{\basemodel{4},\basemodel{4},\basemodel{4}\}   & 47.2 &\ding{51}&\textbf{10.7}& $46.19$ & $46.04$ & ${51.39}$ & ${34.31}$ & $64.00$ & $39.16$ & ${48.23}$ & ${48.39}$ \\
        
         &&Ours: \network{ReVT} \{\basemodel{4},\basemodel{5},\basemodel{6}\}   & 47.2 &\ding{51}&\textbf{10.7}&$\underline{47.95}$ & $\mathbf{48.26}$&$\mathbf{52.59}$ & $\mathbf{36.80}$ & $64.70$ & $\underline{40.96}$ & $\mathbf{49.84}$ & $\underline{50.06}$ \\\
         &&Ours: \network{ReVT} \{\basemodel{1},\basemodel{4},\basemodel{6}\}   & 47.2 &\ding{51}&\textbf{10.7}&$\mathbf{48.33}$ & $\underline{48.17}$ & $\underline{52.28}$ & $\underline{36.67}$ & ${65.14}$ & $\mathbf{41.38}$ & $\underline{49.74}$ & $\mathbf{50.12}$ \\
        \midrule[.9pt]
         \parbox[t]{2mm}{\multirow{4}{*}{\rotatebox[origin=c]{90}{\small{{\textbf{Group 3}}}}}} &\parbox[t]{2mm}{\multirow{4}{*}{\rotatebox[origin=c]{90}{\small{\network{MiT-B5}}}}} &Baseline & 84.7&\ding{51} &\textbf{9.7}& $45.31$ & $43.32$ & $46.85$ & $33.81$ & $\mathbf{68.17}$ & $36.22$ & $46.16$ & $47.49$ \\
       
          &&Ours: \network{ReVT} \{\raisebox{.8pt}{\textcircled{\raisebox{-0.9pt}{4}}},\raisebox{.8pt}{\textcircled{\raisebox{-0.9pt}{4}}},\raisebox{.8pt}{\textcircled{\raisebox{-0.9pt}{4}}}\}   & 84.7 &\ding{51}&\textbf{9.7}& $48.81$ & $47.52$ & $\mathbf{53.21}$ & $36.18$ & $65.67$ & $39.19$ & $45.86$ & $50.28$ \\
         & &Ours: \network{ReVT} \{\basemodel{4},\basemodel{5},\basemodel{6}\}   & 84.7 &\ding{51}&\textbf{9.7}&$\underline{49.55}$ & $\mathbf{48.11}$ & $\underline{53.06}$ & $\underline{36.86}$ & $66.38$ & $\underline{40.36}$ & $\underline{46.88}$ & $\underline{50.79}$ \\
         & &Ours: \network{ReVT} \{\raisebox{.8pt}{\textcircled{\raisebox{-0.9pt}{1}}},\raisebox{.8pt}{\textcircled{\raisebox{-0.9pt}{4}}},\raisebox{.8pt}{\textcircled{\raisebox{-0.9pt}{6}}}\}   & 84.7 &\ding{51}&\textbf{9.7}& $\mathbf{49.96}$ & $\underline{48.01}$ & $52.76$ & $\mathbf{37.27}$ & $\underline{66.40}$ & $\mathbf{41.15}$ & $\mathbf{50.39}$ & $\mathbf{50.88}$ \\
        
        \bottomrule[.9pt]
    \end{tabular}

  \label{tab:sota} 
  \vspace{-4.5mm}
\end{table*}
It can be seen that \textit{we exceed the performance of prior work that is comparable in network size}. In group 1, our \network{ReVT} with the \network{MiT-B2} encoder achieves a benchmark mIoU (BM mean) of $47.29\%$, excelling the best prior work (WildNet with \network{ResNet-50}, 46.33\%), while having fewer parameters ($27.4$ M vs.\ $43.7$ M parameters) and a higher framerate (12 fps vs.\ 7.9 fps). In group 2, the \network{ReVT} with the \network{MiT-B3} achieves a BM mean performance of $50.12\%$, which is $2.31\%$ abs.\ higher than the best prior work (WildNet with \network{ResNet-101}, 47.81\%), while having fewer parameters ($47.2$ M vs.\ $62.7$ M) and a higher frame rate (10.7 fps vs.\ 5 fps). In both groups, our method performs slightly worse in the source domain (GTA5), which is included in the BM mean, which, however, has little relevance for practical real-world applications. \\ \indent\textit{It should also be noted, that our method does not employ any real auxiliary domains for image stylization such as WildNet}~\cite{Lee2022wildnet}\textit{, DRPC}{\cite{Yue2019} \textit{and} \textit{FSDR}~\cite{Huang2021}.
It can also be seen that our largest \network{ReVT} with an \network{MiT-B5} encoder (group 3) achieves the overall highest performance of all evaluated models with a BM mean mIoU of $50.88\%$, still having a higher frame rate than \network{ResNet-50}-based WildNet~\cite{Lee2022wildnet}, which achieves only a BM mean of 46.33\%. \\
\indent The higher mIoU values on the additional target domains (ACDC and KITTI) further indicate the excellent generalization capability of the \network{ReVTs}.
\section{Conclusions}
In this work we show how to improve the domain generalization capabilities of a vision transformer for semantic segmentation with a simple but effective augmentation and re-parameterization method (\network{ReVT}). We show the effect of different image augmentations and optimizer methods on the re-parameterization. Our method is smaller and computationally more efficient than network and encoder ensembles and also achieves state-of-the-art performance in the synthetic-to-real domain generalization task for semantic segmentation, exceeding prior art. In contrast to some prior art, our \network{ReVT} does not require an additional real auxiliary domain during training. We achieve a top mean mIoU of $50.88\%$, when using the largest model and also improve on the best prior art by $0.96\%$ and $2.31\%$ absolute using models with fewer parameters and a higher~frame~rate.
{\small
\bibliographystyle{ieee_fullname}
\bibliography{ifn_spaml_bibliography}
}
\clearpage
\newpage
\clearpage
\section*{Supplementary Material}
In this supplementary material we give a detailed overview of the training and evaluation settings along with hyperparameters. We also provide a detailed description of the employed augmentation methods. Further, we show additional ablation studies and investigations. We also depict the \network{SegFormer} architecture in a block diagram. Finally, we discuss limitations and ethical implications for our method. 
\renewcommand{\thesubsection}{\Alph{subsection}}
\subsection{Detailed Description of Augmentation Methods}
\label{supp:augmenations}
In the following section, we present a detailed
description of the employed augmentation methods. \par
\textbf{Random crop}: Parameter $\varrho$ defines the maximum proportion a single class can occupy in the random crop.\par
\textbf{PhotoAug}:
Photometric augmentation (PhotoAug) comprises the following steps: Each transformation is applied to the image with a probability of 0.5. The position of the random contrast adjustment is in second (mode \ding{202}) or second to last position (mode \ding{203}). The position is randomly selected for each image.
\begin{enumerate}
\setlength{\itemsep}{0pt}
\setlength{\parskip}{0pt}
  \item random brightness
  \item if \ding{202}: random contrast 
  \item convert color from RGB to HSV 
  \item random saturation 
  \item random hue 
  \item convert color from HSV to RGB
  \item if \ding{203}: random contrast 
  \item randomly swap channels 
\end{enumerate}
\par\textbf{Bilateral filter}:
A bilateral filter smoothes an image while preserving sharp edges. Its focus is on the removal of noise and textures. In general, it is a Gaussian filter that smoothes less in non-uniform regions (edge regions) and more in uniform image regions (non-edge regions). The bilateral filter at pixel index $i$ can be described as:
\begin{equation}
  G(i) = \frac{1}{w}\sum_{j\in\mathcal{J}}{N}(\Delta_{ij};\sigma_s^2){N}(||\VEC{x}_i-\VEC{x}_j||_2;\sigma_c^2)\VEC{x}_j \ ,
\end{equation} with 
\begin{equation}
N(d;\sigma^2)=e^{-\frac{1}{2}(\frac{d}{\sigma})^2} \ ,
\end{equation}
and normalizing factor
\begin{equation}
  w = \sum_{j\in\mathcal{J}}{N}(\Delta_{ij};\sigma_s^2){N}(||\VEC{x}_i-\VEC{x}_j||_2;\sigma_c^2).
\end{equation}
The neighboring pixel index is denoted as $j$ and stems from the neighborhood $\mathcal{J}$. The neighborhood is defined by the kernel size which we sample from a uniform distribution between 1 (px) and 15 (px). Distances $\Delta_{ij} = \sqrt{|h_i-h_j|^2+|w_i-w_j|^2}$ and $||\VEC{x}_i-\VEC{x}_j||_2$ denote the (Euclidean) pixel distance and color difference, respectively, where $h_i$ and $w_i$ are the height and width position of pixel $i$ and $\VEC{x}_i \in \mathbb{G}^3$ denotes the vector of RGB values for pixel $i$ (likewise for pixel $j$). We set the spatial distance to $\sigma_s\!=\!75$ and the color distance to $\sigma_c\!=\!75$. The probability for applying this filter is set to $p\!=\!0.5$.
\par\textbf{PixMix}:
The PixMix~\cite{Hendrycks2022pixmix} augmentation method comprises multiple processing steps as shown in \autoref{lst:pixmix}. Here, $\VEC{x}$ denotes the input image and $\VEC{z}\in\mathcal{Z}$ denotes the mixing image from the PixMix set of fractal images $\mathcal{Z}$~\cite{Hendrycks2022pixmix}. We set the maximum number of mixing rounds to $K\!=\!3$. Note that the for loop is not executed for random choice = 0. The \textbf{mix\_op($\cdot$)} function is randomly chosen to be either addition (\textbf{add}) or multiplication (\textbf{multiply}). It gets the images $\tilde{\VEC{x}}_{k-1}$ and $\tilde{\VEC{x}}_\mathrm{mix}$ as inputs, as well as $\beta\!=\!3$, which is used to generate independent weighting factors for the images. The weighting factors are sampled from a Beta distribution.
For the \textbf{augment($\cdot$)} function in the PixMix pseudocode we used only baseline augmentation methods PhotoAug and Random Flip (cf.\ \autoref{fig:00-augmenation-pipeline}), that is why we denote the method as PixMix*. 

\begin{lstlisting}[float=t,language=Python, caption=\textbf{Pseudo-code} of the \textbf{PixMix}~\cite{Hendrycks2022pixmix} data augmenation.,escapeinside={(*}{*)},morekeywords={mix_op,add,multiply,augment},label={lst:pixmix}]
def PixMix((*$\mathbf{x,z},K,\beta$*)): # mixing image (*\color{tu3}$\mathbf{z} \in \mathcal{Z}$*)
    (*$\tilde{\mathbf{x}}_0$*) = random.choice({augment((*$\mathbf{x}$*)),(*$\mathbf{x}$*)})
    # random number of mixing rounds
    for (*k*) = 1:random.choice({0,1,...,(*K*)}):
        (*$\mathbf{x_\mathrm{mix}}$*) = random.choice({augment((*$\mathbf{x}$*)),(*$\mathbf{z}$*)})
        mix_op = random.choice({add,multiply}) 
        (*$\tilde{\mathbf{x}}_k$*) = mix_op((*$\tilde{\mathbf{x}}_{k-1}, \mathbf{x}_\mathrm{mix}, \beta$*))
    return (*$\tilde{\mathbf{x}}_k$*)
\end{lstlisting}

\par
\subsection{Training/Evaluation Settings, Hyperparameters}
\label{supp:settings}
\setcounter{table}{8}
\begin{table*}[t]
  \centering
  \renewcommand{\arraystretch}{.8}
  \setlength{\tabcolsep}{.35em}
  \caption{\textbf{Settings and hyperparameters} for the \network{SegFormer} and \network{DeepLabv3+} training.}

\extrarowheight=\aboverulesep
    \addtolength{\extrarowheight}{\belowrulesep}
    \aboverulesep=0pt
    \belowrulesep=0pt
    \begin{tabular}{@{}l@{\hskip 7mm}ab@{}}
        \toprule
        \makecell[l]{\textbf{Setting / Hyperparameter}}  & \network{SegFormer} & \network{DeepLabv3+} \\
       \midrule 
       Optimizer &AdamW~\cite{Loshchilov2019}&SGD \\
       \# of training iterations ($\tau_\mathrm{max}$) & 40,000& 60,000 \\
       Momentum values (AdamW) ($\beta_1,\beta_2$) & 0.9, 0.999& - \\
       Momentum  ($\beta$)& - & 0.9\\
       Warm-up iterations & 1500 & - \\
       Warm-up ratio & $1\!\cdot\!10^{-6}$ & -\\
       Initial LR ($\eta_0$) &$6\!\cdot\!10^{-5}$&$1\!\cdot\!10^{-3}$ \\
       Weight decay $\omega$ & 0.01& 0.0005 \\
       Learning rate  (LR) schedule ($\eta(\tau)$) & Polynomial (\ref{eq:lr-schedule})  &Polynomial (\ref{eq:lr-schedule})\\

       Batch size &2 & 2 \\
       Random decoder init  & Kaiming initialization & Kaiming initialization\\
         Resized input resolution $\gtavtrain$& $720\!\times\!1280$ & $720\!\times\!1280$ \rule[-1.2ex]{0mm}{3.65ex}  \\
       
        \bottomrule
    \end{tabular}

  \label{tab:hyperparameters} 
\end{table*}
\setcounter{table}{7}
\begin{table}[h]
  \centering
  \renewcommand{\arraystretch}{.8}
  \setlength{\tabcolsep}{.25em}
  \caption{\textbf{Image and label resolution [$\mathrm{px}\!\times\!\mathrm{px}$]} for the employed \textbf{evaluation} datasets. $^\circ$In both GTA and KITTI, there are images that differ by a few pixels from their normal resolution. *The crowd-sourced Mapillary Vistas dataset does not have a fixed resolution, but a highly variable one.}

\extrarowheight=\aboverulesep
    \addtolength{\extrarowheight}{\belowrulesep}
    \aboverulesep=0pt
    \belowrulesep=0pt
    \begin{tabular}{@{}lcc@{}}
        \toprule
         \textbf{Dataset name}& \multicolumn{2}{c}{\textbf{Resolution ($H\!\times\!W$) of ...}} \\
        & \textbf{\makecell{resized images}} & \textbf{\makecell{labels}}\\
        \midrule
        GTA5~\cite{Richter2016}&    $512\!\times\!\phantom{1}932^\circ$ &$1052\!\times\!1914^\circ$ \\
        SYNTHIA~\cite{Ros2016} (SYN) &   $512\!\times\!\phantom{1}862\phantom{^*}$ &$\phantom{1}760\!\times\!1280\phantom{^*}$ \\
        \midrule
        Cityscapes~\cite{Cordts2016} (CS)&  $512\!\times\!1024\phantom{^*}$ &$1024\!\times\!2048\phantom{^*}$ \\
        Mapillary Vistas~\cite{Neuhold2017} (MV) &  various${^*}$  &various${^*}$ \\
        BDD100k~\cite{Yu2018b} (BDD) &    $512\!\times\!\phantom{1}910\phantom{^*}$ &$\phantom{1}720\!\times\!1280\phantom{^*}$ \\
        ACDC~\cite{Sakaridis2021acdc} &    $512\!\times\!\phantom{1}910\phantom{^*}$ &$1080\!\times\!1920\phantom{^*}$ \\
        KITTI~\cite{AbuAlhaija2018} (KIT) & $309\!\times\!1024^\circ$ & $\phantom{1}375\!\times\!1242^\circ$ \\
        \bottomrule
    \end{tabular}

  \label{tab:eval-resolutions} 
\end{table}
In the following section, we will provide a detailed overview of the training and evaluation settings and hyperparameters. For the training and evaluation we employ \network{PyTorch} v.$3.8.13$ and the \network{MMSegmentation} toolbox v.$0.11.0$. Additionally, we refer to our repository, where all code for the conducted experiments is made available\footnote{Code is available at \href{https://github.com/ifnspaml/ReVT}{https://github.com/ifnspaml/ReVT}}. \par
\textbf{Training phase}:
In \autoref{tab:hyperparameters} we list all settings and hyperparameters that were used for the training process.
The polynomial learning rate schedule is defined as follows:
\begin{equation}
  \label{eq:lr-schedule}
  \eta(\tau) = \eta_0 (1-\frac{\tau}{\tau_{\mathrm{max}}})^{0.9},
\end{equation}
with $\eta(\tau)$ being the learning rate at optimizer step (iteration) $\tau$ and $\eta_0$ being the initial learning rate. The maximum number of iterations is given by $\tau_{\mathrm{max}}$. \\
During training, the images from the source domain $\src$ get resized to a resolution of $720\!\times\!1280$. \par
\textbf{Evaluation phase}:
For evaluation, we always employ the final model weights after the full training and do not perform any checkpoint selection. We resize the input images during evaluation in a way that the image will be rescaled as large as possible within a pre-defined scale ($512\!\times\!1024$), while still keeping their aspect ratios. The network output is then resized to the original image resolution. The mIoU is calculated on the original resolution, also referred to as label resolution. The resized image and the original label resolutions for all employed datasets are listed in \autoref{tab:eval-resolutions}. \par The frame rate computations were performed on the rescaled Cityscapes dataset. We used 200  images for inference and computed the mean frame rate after a warmup phase of five images to account for any delays due to image reading operations. 
\subsection{Additional Details on the Choice of Optimizer}
\label{supp:add-optimizer-exp}
In our experiments in \autoref{tab:optimizer} we show that the choice of optimizer has a strong effect on the baseline performance of the models, as well as on the performance after re-parameterization. In \autoref{fig:reparam-during-training}, we compare the \network{SegFormer} architecture with its standard optimizer setup (left) and the \network{DeepLabv3+} architecture with its standard optimizer setup (right). We show the mean cosine similarity between three baseline (\basemodel{1}) models for the encoder only ($\VECG{\theta}^\mathrm{E}$, upper plots) and the full model ($\VECG{\theta}$, center plots), during the training process. Note that the standard number of iterations differs for both models and is 40,000 for the \network{SegFormer} and 60,000 for the \network{DeepLabv3+}. \par 
It can be seen that the mean cosine similarity for the network parts that are re-parameterized in our method (encoder only) have a similar mean cosine similarity after the training for both networks (0.995). In the bottom plots of the figure, we report the mIoU values for both in-domain (GTA5, green) and out-of-domain (OOD) data (Cityscapes, red) for the baseline (dashed lines) and re-parameterized (solid lines) models. It can be seen that the performance of the re-parameterized models is higher for the \network{SegFormer} for any training iteration. For the \network{DeepLabv3+}, however, the baseline performance is always higher for in-domain data (green) and fluctuates for OOD data (red), but ultimately the baseline performance is also higher for OOD data in the last iterations.
\begin{figure*}[t]
\centering\includegraphics{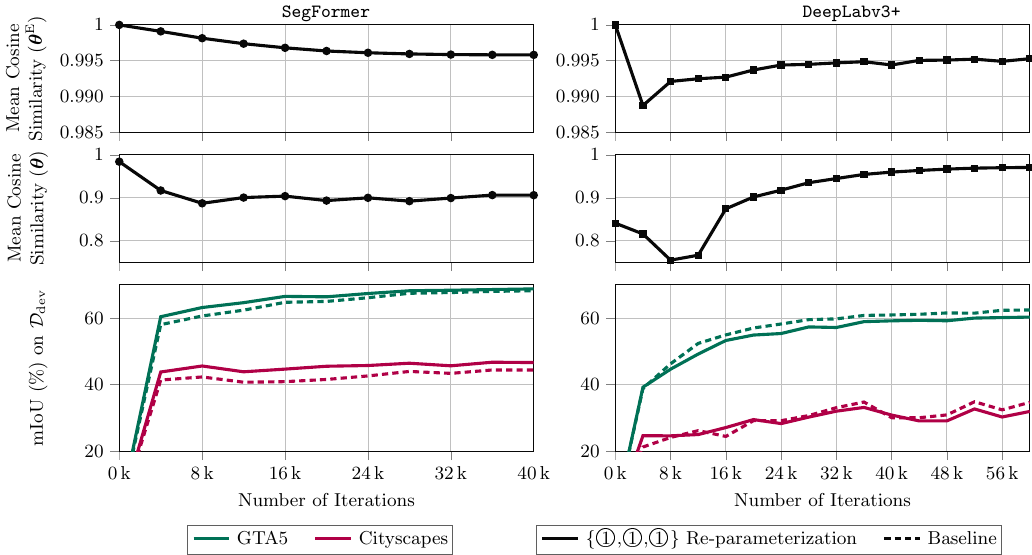}
  \caption{Comparison of \textbf{mean cosine similarity vs.\ mIoU performance} for \network{SegFormer} and \network{DeepLabv3+} \textbf{during training}. The first row shows the mean cosine similarity for the encoder only ($\VECG{\theta}^\mathrm{E}$), the second row for the full network ($\VECG{\theta}^\mathrm{E}$). The mean cosine similarity is computed between three models. In the bottom row, the mIoU is given for the dev sets of GTA5 (\textcolor{tu10}{green}) and Cityscapes (\textcolor{tu3}{red}).}
  \label{fig:reparam-during-training}
\end{figure*}
\par Since the mean cosine similarity did not provide any insights into the causes for the poor performance of the re-parameterized \network{DeepLabv3+}, we further investigated the mean cosine similarity for individual layers $\ell$ of the networks as shown in~\autoref{fig:reparam-vs-optimizer}.
\definecolor{myyellow}{rgb}{1,0.7,0.0}
\begin{figure}[t]
\centering\includegraphics[width=\columnwidth]{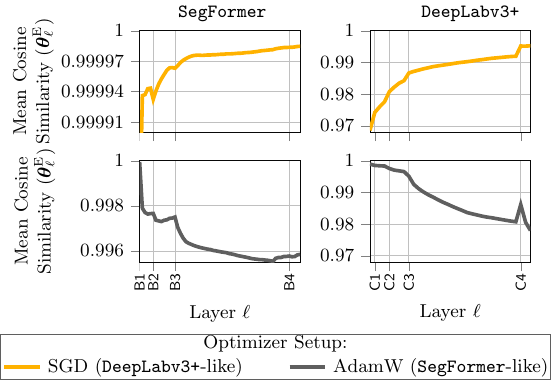}
  \caption{Comparison of the \textbf{layer-wise mean cosine similarity} for the encoder only ($\VECG{\theta}^\mathrm{E}_\ell$). Shown are results for the \network{SegFormer} (left) and \network{DeepLabv3+} (right) architectures, which were trained with the standard \network{DeepLabv3+} or \network{SegFormer} optimizer setup, shown in {gray} and {yellow}, respectively. The mean cosine similarity is computed between three models. For the purpose of clarity, we only indicate the individual network blocks (\textsf{B}$b$ for \network{SegFormer}, \textsf{C}$c$ for \network{DeepLabv3+}) on the x axis.}
  \label{fig:reparam-vs-optimizer}
\end{figure}
We show the layer-wise mean cosine similarity for the encoder network ($\VECG{\theta}^\mathrm{E}_\ell$), where $\ell$ indicates the layer index. For the purpose of clarity, we only mark the first layer of each of the major network blocks. For the \network{SegFormer}, we indicate the transformer blocks by \textsf{B}$b$ with $b$ being the block index (cf.\ \autoref{fig:blockdiagramm-mitb5}). For the \network{DeepLabv3+}, we indicate the convolutional blocks, as defined by He et al.\ [\textcolor{green}{48}], by \textsf{C}$c$ with $c$ being the block index. \textit{It can be seen that the choice of optimizer has a significant effect on the layer-wise cosine similarity.} For the standard \network{DeepLabv3+} optimizer setup with SGD, the cosine similarity for both network architectures is higher in deeper layers and lower in earlier layers. In contrast, when the standard \network{SegFormer} optimizer setup with AdamW is employed, the cosine similarity is highest for earlier layers and drops for deeper layers. This specific property might be important for a well-performing encoder re-parameterization. As already shown in \autoref{tab:optimizer}, this optimizer setup (AdamW~\cite{Loshchilov2019}) also allows the \network{DeepLabv3+} to improve over its baseline performance. Accordingly, we used the AdamW optimizer for our \network{ReVT} method in the main paper.
\subsection{Additional Ablation Studies}
In this section, we will investigate the weighting of the base models and compare the \network{ReVT}
re-parameterization w.r.t. re-parameterized network parts,
layers, and the number of base models. Afterwards, we
evaluate different base model augmentations and optimizer
methods during training to design our final ReVT. 
\label{supp:add-experiments}
\par\textbf{Weighting of networks}:
In \autoref{fig:/model-reparametrization-ternary} we depict multiple possible weighting combinations for three models with the best combination (marked with a blue circle) achieving an mIoU of $47.68\%$, while the uniform re-parameterization (all models weighted with $\frac{1}{3}$, marked with a yellow circle) achieves an mIoU of $47.49\%$. It can be seen that the weighting of the individual models is actually quite insensitive, which is why we decide to use the simple variant of the uniform encoder re-parameterization (i.e., weights $\frac{1}{3}$). 
\begin{figure}[t]
  \centering
  \includegraphics[width=\columnwidth]{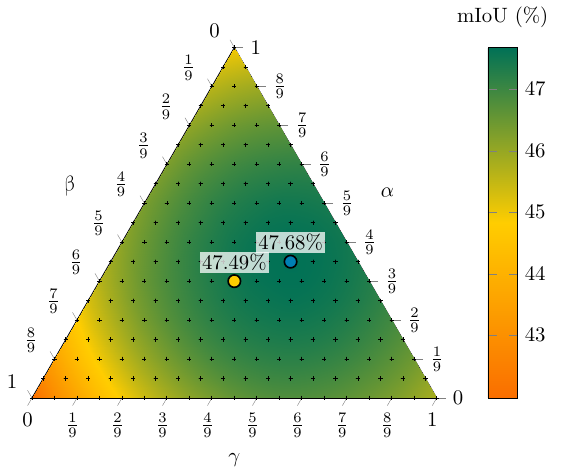}
  \caption{Ternay plot showing the performance (mIoU (\%)) of three baseline models \{\basemodel{1},\basemodel{1},\basemodel{1}\} and multiple weight combinations ($\alpha,\beta,\gamma$). The mIoU is calculated by a re-parameterization, where the parameters of the three models are weighted by the values of $\alpha$, $\beta$, and $\gamma$, respectively. The \textbf{training} of the base models (\network{SegFormer}) was performed on the synthetic \textbf{GTA5} training set ($\src\!=\!\gtavtrain$). The \textbf{evaluation} is performed on the \textbf{Cityscapes development set} ($\tgt\!=\!\csdev$). The center of the plot ($\alpha\!=\!\beta\!=\!\gamma\!=\!\frac{1}{3}$) corresponds to a uniform encoder re-parameterization.}
  \label{fig:/model-reparametrization-ternary}
\end{figure}
\par\textbf{\network{ReVT} vs.\ ensembles}:
In \autoref{tab:ensemble-vs-revt} we compare the \network{ReVT} and a network ensemble for different combinations of training settings. It can be seen that the \network{ReVT} method outperforms the network ensemble not only for the combination of three baseline models \{\basemodel{1},\basemodel{1},\basemodel{1}\}, as already shown in \autoref{fig:numofmodels}, but also for all other tested base model combinations $\{\raisebox{2pt}{\Circled[inner ysep=3.5pt,inner xsep=0pt]{$a_{\scaleto{1}{4pt}}$}},\raisebox{2pt}{\Circled[inner ysep=3.5pt,inner xsep=0pt]{$a_{\scaleto{2}{4pt}}$}},\raisebox{2pt}{\Circled[inner ysep=3.5pt,inner xsep=0pt]{$a_{\scaleto{3}{4pt}}$}}\}$ on the real test$*$ sets. However, the network ensemble is slightly better in the synthetic source domain ($\gtavdev$), which has little relevance for real-world applications. The same applies to the base model combination \{\basemodel{5},\basemodel{5},\basemodel{5}\}, where the ensemble performs slightly better on the SYNTHIA dataset ($\synthiadev$). The overall best performance for each dataset is highlighted in light green. It can be seen that the \network{ReVT} method achieves top performance for each dataset and is even on par with the ensemble on the source dataset ($\gtavdev$). In general, once again, the \network{ReVT} \{\basemodel{1},\basemodel{4},\basemodel{6}\} seems to be a strong \network{ReVT}, generalizing well to unseen datasets.
\setcounter{table}{9}
\begin{table}[t]
  \centering
  \renewcommand{\arraystretch}{0.8}
  \setlength{\tabcolsep}{.16em}
  \caption{Performance (mIoU (\%)) of a \textbf{network ensemble} vs.\ the \textbf{\network{ReVT}} for different base model combinations. The \textbf{training} of the base models (\network{SegFormer}) was performed on the \textbf{GTA5} ($\src\!=\!\gtavtrain$) dataset. The \textbf{evaluation} is performed on the \textbf{GTA5} and \textbf{SYNTHIA development sets} and on the \textbf{test$*$ data} of various real-world target datasets ($\tgt\!=\!\datateststar$). Reported is the mean value over the respective number of employed base models. Best result for each each base model combination in bold face, overall best performance per dataset is highlighted in \textcolor{tu9}{light green}.}
  
\definecolor{mygreen}{rgb}{0.153,0.44,0.0}
\extrarowheight=\aboverulesep
    \addtolength{\extrarowheight}{\belowrulesep}
    \aboverulesep=0pt
    \belowrulesep=0pt
    \begin{tabular}{@{}ccccccc@{}}
        \toprule
         \makecell[c]{\textbf{Base} \\ \textbf{Models} }& \textbf{Method} &    \multicolumn{5}{c}{\textbf{mIoU (\%) on}} \\
      \cmidrule{3-7}
        && $\gtavdev$ & $\synthiadev$ & $\csteststar$ & $\bddteststar$ &$\mvteststar$ \rule[-1.1ex]{0mm}{3.55ex} \\
       \midrule      
        \multirow{2}{*}{\{\basemodel{1},\basemodel{1},\basemodel{1}\}} & Ensemble & $\mathbf{68.8}$ & ${34.6}$ & ${46.9}$ & ${44.8}$ & ${48.0}$ \\
         & \texttt{ReVT} & ${68.6}$ & $\mathbf{35.5}$ & $\mathbf{49.3}$ & $\mathbf{45.3}$ & $\mathbf{49.3}$ \\
        \midrule
        \multirow{2}{*}{\{\basemodel{2},\basemodel{2},\basemodel{2}\}} & Ensemble & $\mathbf{69.2}$ & ${33.1}$ & ${43.8}$ & ${43.9}$ & ${46.6}$ \\
         & \texttt{ReVT} & ${69.1}$ & $\mathbf{34.1}$ & $\mathbf{44.9}$ & $\mathbf{44.1}$ & $\mathbf{47.6}$ \\
          \midrule
        \multirow{2}{*}{\{\basemodel{3},\basemodel{3},\basemodel{3}\}} & Ensemble & $ \cellcolor{tu92}{\mathbf{69.7}}$ & ${33.8}$ & ${43.8}$ & ${42.6}$ & ${47.3}$ \\
         & \texttt{ReVT} & $ \cellcolor{tu92}{\mathbf{69.7}}$ & $\mathbf{35.0}$ & $\mathbf{45.7}$ & $\mathbf{43.5}$ & $\mathbf{48.8}$ \\
          \midrule
        \multirow{2}{*}{\{\basemodel{4},\basemodel{4},\basemodel{4}\}} & Ensemble & $\mathbf{65.9}$ & $\mathbf{36.8}$ & ${47.5}$ & ${47.4}$ & ${52.6}$ \\
         & \texttt{ReVT} & ${65.7}$ & ${36.2}$ & $\mathbf{48.8}$ & $\mathbf{47.5}$ & $\mathbf{53.2}$ \\
          \midrule
        \multirow{2}{*}{\{\basemodel{5},\basemodel{5},\basemodel{5}\}} & Ensemble & $\mathbf{68.6}$ & ${35.1}$ & ${47.5}$ & ${45.4}$ & ${50.4}$ \\
         & \texttt{ReVT} & $\mathbf{68.5}$ & $\mathbf{35.9}$ & $\mathbf{48.6}$ & $\mathbf{45.9}$ & $\mathbf{51.2}$ \\
          \midrule
        \multirow{2}{*}{\{\basemodel{6},\basemodel{6},\basemodel{6}\}} & Ensemble & $\mathbf{65.0}$ & $\mathbf{37.0}$ & ${48.0}$ & ${47.9}$ & ${52.8}$ \\
         & \texttt{ReVT} & ${64.9}$ & ${36.1}$ & $\mathbf{49.7}$ & $ \cellcolor{tu92}{\mathbf{48.5}}$ & $ \cellcolor{tu92}{\mathbf{53.5}}$ \\
          \midrule
        \multirow{2}{*}{\{\basemodel{4},\basemodel{5},\basemodel{6}\}} & Ensemble & $\mathbf{67.2}$ & ${35.1}$ & ${48.1}$ & ${47.3}$ & ${51.6}$ \\
         & \texttt{ReVT} & $\mathbf{66.4}$ & $\mathbf{36.9}$ & $\mathbf{49.5}$ & $\mathbf{48.1}$ & $\mathbf{53.1}$ \\
          \midrule
        \multirow{2}{*}{\{\basemodel{1},\basemodel{4},\basemodel{6}\}} & Ensemble & $\mathbf{68.5}$ & ${34.2}$ & ${47.0}$ & ${45.9}$ & ${50.1}$ \\
         & \texttt{ReVT} & ${66.4}$ & $ \cellcolor{tu92}{\mathbf{37.3}}$ & $ \cellcolor{tu92}{\mathbf{50.0}}$ & $\mathbf{48.0}$ & $\mathbf{52.8}$ \\
        \bottomrule
    \end{tabular}

  \label{tab:ensemble-vs-revt} 
\end{table}
\subsection{\textbf{\network{ReVT}} with SYNTHIA as Source}
In this section, we provide additional results for our method, when trained with the SYNTHIA dataset as source domain: $\src\!=\!\synthiatrain$. 
In \autoref{tab:method-design-synthia} we evaluate the various augmentation methods \basemodel{\raisebox{1pt}{$a$}} that were already evaluated for models trained on GTA5 ($\src\!=\!\synthiatrain$) in \textcolor{red}{Section}~\ref{sec:experiments}. In the lower part of the table we report some ($M\!=\!3$) combinations $\{\raisebox{2pt}{\Circled[inner ysep=3.5pt,inner xsep=0pt]{$a_{\scaleto{1}{4pt}}$}},\raisebox{2pt}{\Circled[inner ysep=3.5pt,inner xsep=0pt]{$a_{\scaleto{2}{4pt}}$}},\raisebox{2pt}{\Circled[inner ysep=3.5pt,inner xsep=0pt]{$a_{\scaleto{3}{4pt}}$}}\}$ of these base models by our re-parameterization. The gray columns indicate our development sets ($\datadev$), where the light grey column is $\synthiadev$. The OOD mean mIoU of $\gtavdev$ and $\csdev$ is shown in the dark grey columns. 
\begin{table*}[t]
  \centering
  \renewcommand{\arraystretch}{0.9}
  \setlength{\tabcolsep}{.2em}
  \caption{Performance (mIoU (\%)) of the \network{SegFormer} model (with an \network{MiT-B5} encoder) using different domain generalization methods. \textbf{Training} was performed on the synthetic \textbf{SYNTHIA} ($\src\!=\!\synthiatrain$) dataset. \textbf{Evaluation} is performed on the \textbf{Cityscapes}, \textbf{GTA5}, and \textbf{SYNTHIA development sets} (gray columns) and on the \textbf{test$*$ data} of various real-world target datasets ($\tgt\!=\!\datateststar$). 
Reported is the mean mIoU $\pm$ the standard deviation of $M\!=\!3$ models with various image augmentations. For the \network{ReVT}, the mean $\pm$ standard deviation is computed with one averaged encoder and the three associated decoders $m\in\{1,2,3\}$. For models trained on SYNTHIA, we evaluate over 16 classes, as is common practice~\cite{Klingner2020c}. Best results in bold face, second-best underlined.
  }

\extrarowheight=\aboverulesep
    \addtolength{\extrarowheight}{\belowrulesep}
    \aboverulesep=0pt
    \belowrulesep=0pt
    \resizebox{\textwidth}{!}{%
    \begin{tabular}{@{}clhgggcccc@{}}
        \toprule
        \stzdown \hphantom{W} & \makecell[l]{\textbf{Method} \\ performed:} &    \multicolumn{8}{c}{\textbf{mIoU (\%) on}}  \\
      \cmidrule{3-10}
        &&$\synthiadev$ & $\gtavdev$ & $\csdev$ & \raisebox{-3.2pt}{\shortstack{\textbf{OOD} \\\textbf{mean}}} & $\csteststar$ & $\bddteststar$ &$\mvteststar$  &\raisebox{-3.0pt}{\shortstack{\textbf{test$*$} \\\textbf{mean}}}  \rule[-1.1ex]{0mm}{3.55ex}  \\
       \midrule
       \parbox[t]{2mm}{\multirow{6}{*}{\rotatebox[origin=c]{90}{... during training}}} &Baseline \basemodel{1}& $76.5 \!\pm\! 0.1$ & $42.8 \!\pm\! 0.4$ & $\mathbf{44.3} \!\pm\! 1.3$ & $\mathbf{43.5} \!\pm\! 1.2$ & $\mathbf{45.1} \!\pm\! 1.6$ & $35.2 \!\pm\! 1.4$ & $\mathbf{42.5} \!\pm\! 0.9$ & $\underline{40.9} \!\pm\! 4.4$ \\

       &  \minus PhotoAug \basemodel{2}& $\underline{77.3} \!\pm\! 0.1$ & $39.8 \!\pm\! 0.6$ & $41.4 \!\pm\! 0.5$ & $40.6 \!\pm\! 0.9$ & $41.6 \!\pm\! 0.8$ & $33.7 \!\pm\! 1.0$ & $40.8 \!\pm\! 0.6$ & $38.7 \!\pm\! 3.6$ \\
    
       &  \minus PhotoAug, \minus Rand. Flip \basemodel{3}&$\mathbf{78.3} \!\pm\! 0.0$ & $40.7 \!\pm\! 0.7$ & $41.3 \!\pm\! 0.3$ & $41.0 \!\pm\! 0.6$ & $41.8 \!\pm\! 0.2$ & $34.3 \!\pm\! 1.1$ & $41.3 \!\pm\! 0.3$ & $39.1 \!\pm\! 3.5$ \\
       
       & +PixMix*~\cite{Hendrycks2022pixmix} \basemodel{4}&$73.8 \!\pm\! 0.1$ & $\mathbf{43.1} \!\pm\! 1.0$ & $42.6 \!\pm\! 1.7$ & $42.8 \!\pm\! 1.4$ & $42.6 \!\pm\! 2.2$ & $\mathbf{38.4} \!\pm\! 0.8$ & $\mathbf{42.5} \!\pm\! 0.6$ & $\mathbf{41.2} \!\pm\! 2.4$ \\
       
       & +Bilateral Filter (BF)~\cite{Tomasi1998bilateral}\basemodel{5} &$76.0 \!\pm\! 0.1$ & $\underline{42.9} \!\pm\! 0.9$ & $43.2 \!\pm\! 0.7$ & $\underline{43.1} \!\pm\! 0.8$ & $\underline{43.1} \!\pm\! 0.9$ & $36.0 \!\pm\! 0.7$ & $41.7 \!\pm\! 0.4$ & $40.3 \!\pm\! 3.1$ \\
       & +PixMix*~\cite{Hendrycks2022pixmix} +BF~\cite{Tomasi1998bilateral} \basemodel{6}&$72.3 \!\pm\! 0.1$ & $41.2 \!\pm\! 0.6$ & $\underline{43.3} \!\pm\! 0.4$ & $42.2 \!\pm\! 1.2$ & $43.0 \!\pm\! 0.5$ & $\underline{38.0} \!\pm\! 0.6$ & $41.6 \!\pm\! 0.3$ & $\underline{40.9} \!\pm\! 2.2$ \\   
       \midrule
      \parbox[t]{2mm}{\multirow{8}{*}{\rotatebox[origin=c]{90}{...after training}}} & \network{ReVT} \{\basemodel{1},\basemodel{1},\basemodel{1}\}&$76.2 \!\pm\! 0.1$ & $42.3 \!\pm\! 0.1$ & $44.9 \!\pm\! 0.4$ & $43.6 \!\pm\! 1.3$ & $\underline{45.8} \!\pm\! 0.4$ & $35.8 \!\pm\! 0.2$ & $43.6 \!\pm\! 0.2$ & $41.7 \!\pm\! 4.3$ \\
       & \network{ReVT} \{\basemodel{2},\basemodel{2},\basemodel{2}\}&$\underline{76.9} \!\pm\! 0.0$ & $40.4 \!\pm\! 0.2$ & $42.2 \!\pm\! 0.2$ & $41.3 \!\pm\! 1.0$ & $42.6 \!\pm\! 0.1$ & $34.7 \!\pm\! 0.3$ & $42.0 \!\pm\! 0.3$ & $39.8 \!\pm\! 3.6$ \\
        & \network{ReVT} \{\basemodel{3},\basemodel{3},\basemodel{3}\}&$\mathbf{78.0} \!\pm\! 0.1$ & $41.3 \!\pm\! 0.6$ & $41.9 \!\pm\! 0.3$ & $41.6 \!\pm\! 0.6$ & $42.5 \!\pm\! 0.4$ & $35.2 \!\pm\! 0.3$ & $42.7 \!\pm\! 0.3$ & $40.1 \!\pm\! 3.5$ \\
       & \network{ReVT} \{\basemodel{4},\basemodel{4},\basemodel{4}\}&$73.7 \!\pm\! 0.1$ & $\underline{43.4} \!\pm\! 0.1$ & $44.1 \!\pm\! 0.3$ & $43.7 \!\pm\! 0.4$ & $44.3 \!\pm\! 0.3$ & $39.5 \!\pm\! 0.3$ & $\underline{44.0} \!\pm\! 0.3$ & $42.6 \!\pm\! 2.2$ \\
       & \network{ReVT} \{\basemodel{5},\basemodel{5},\basemodel{5}\}&$75.7 \!\pm\! 0.0$ & $\mathbf{44.1} \!\pm\! 0.7$ & $44.4 \!\pm\! 0.3$ & $\mathbf{44.2} \!\pm\! 0.6$ & $44.5 \!\pm\! 0.3$ & $37.1 \!\pm\! 0.3$ & $43.0 \!\pm\! 0.2$ & $41.5 \!\pm\! 3.2$ \\
       & \network{ReVT} \{\basemodel{6},\basemodel{6},\basemodel{6}\}&$72.2 \!\pm\! 0.1$ & $42.4 \!\pm\! 0.2$ & $44.4 \!\pm\! 0.2$ & $43.4 \!\pm\! 1.0$ & $44.3 \!\pm\! 0.2$ & $38.9 \!\pm\! 0.2$ & $42.9 \!\pm\! 0.0$ & $42.0 \!\pm\! 2.3$ \\
        & \network{ReVT} \{\basemodel{4},\basemodel{5},\basemodel{6}\}&$74.0 \!\pm\! 0.6$ & $43.2 \!\pm\! 0.4$ & $\underline{45.0} \!\pm\! 0.4$ & $\underline{44.1} \!\pm\! 0.9$ & $45.1 \!\pm\! 0.4$ & $\underline{39.6} \!\pm\! 0.4$ & $\underline{44.0} \!\pm\! 0.2$ & $\underline{42.9} \!\pm\! 2.4$ \\
        & \network{ReVT} \{\basemodel{1},\basemodel{4},\basemodel{6}\}&$74.1 \!\pm\! 0.6$ & $42.7 \!\pm\! 0.3$ & $\mathbf{45.7} \!\pm\! 0.5$ & $\mathbf{44.2} \!\pm\! 1.5$ & $\mathbf{46.3} \!\pm\! 0.3$ & $\mathbf{40.3} \!\pm\! 0.5$ & $\mathbf{44.8} \!\pm\! 0.1$ & $\mathbf{43.8} \!\pm\! 2.6$ \\
        \bottomrule
    \end{tabular}}

  \label{tab:method-design-synthia}
  \vspace{-4.5mm}
\end{table*}
It can be seen in the upper part of \autoref{tab:method-design-synthia} that the augmentation methods do not improve the performance as much as for models trained on GTA5. The best OOD mean performance is achieved with the baseline model \basemodel{1}. On the test$*$ mean the PixMix* augmentation works best, followed by the combination of PixMix* and the bilateral filter, and baseline model. \par
\textit{Although the individual augmentations do not perform as well for these models, the} \network{ReVT} \{\basemodel{1},\basemodel{4},\basemodel{6}\}\textit{, which we already identified as our best }\network{ReVT} \textit{in} \textit{\textcolor{red}{Section}~\ref{sec:experiments}, provides both top OOD mean ($44.2\%$) and test$*$ mean ($43.8\%$) performance.} Again, second-best results are achieved with the \network{ReVT} \{\basemodel{4},\basemodel{5},\basemodel{6}\}. Additionally, the \network{ReVT} \{\basemodel{5},\basemodel{5},\basemodel{5}\} is on par with the \network{ReVT} \{\basemodel{1},\basemodel{4},\basemodel{6}\} for the OOD mean performance when trained on SYNTHIA.
\par In the following, we compare also against prior art that have also been evaluated with SYNTHIA as source domain. Again, we choose the \network{ReVT} \{\basemodel{1},\basemodel{4},\basemodel{6}\} and \network{ReVT} \{\basemodel{4},\basemodel{5},\basemodel{6}\} for the comparison with prior art. The results are shown in \autoref{tab:sota-synthia}. In contrast to the models trained on GTA5, for models trained on SYNTHIA, the \network{ReVT} \{\basemodel{1},\basemodel{4},\basemodel{6}\} does not always reach the top BM mean performance. Similar to the GTA5-trained models, the performance of both \network{ReVT} variants remains slightly behind that of the baseline for synthetic source domain data ($\synthiadev$), which, however, has little relevance for practical real-world applications. \par
For the small (group 1) and midsized (group 2) models, the \network{ReVT} \{\basemodel{4},\basemodel{5},\basemodel{6}\} yields a slightly better performance of $45.79\%$ vs.\ $45.44\%$ (baseline: $44.09\%$) and $48.99\%$ vs.\ $48.84\%$ (baseline: $46.72\%$), respectively. For the large models (group 3), the \network{ReVT} \{\basemodel{1},\basemodel{4},\basemodel{6}\} yields the best performance with a BM mIoU of $49.64\%$ (baseline: $48.42\%$). In summary, on the benchmark (BM) data, our proposed SYNTHIA-trained \network{ReVT} models achieve an mIoU improvement of $+1.2\%$ absolute (large models) to $+1.7\%$ absolute (small models).
\par It should be noted that no prior work reported on all datasets necessary for the benchmark (BM) mean when trained with SYNTHIA as source domain. All of our \network{ReVTs} improve on the prior art for the reported domains. Only for the smallest models in group 1 the SAN+SAW method~\cite{Peng2022semanticaware} achieves a higher mIoU on the BDD dataset (best prior art: $35.42\%$ vs.\ ours: $35.18\%$). For the midsized models we already improve on this domain (best prior art: $37.40\%$ vs.\ ours: $38.73\%$), and interestingly significantly excel the SAN+SAW method (ours: $38.73\%$ vs.\ SAN+SAW: $35.98\%$).
\begin{table*}[t] 
  \centering
  \renewcommand{\arraystretch}{.9}
  \setlength{\tabcolsep}{.3em}
  \caption{Performance (mIoU (\%)) of various domain generalization methods employing different segmentation networks, sorted into three performance groups. \textbf{Training} was performed on the synthetic \textbf{SYNTHIA} ($\src\!=\!\synthiatrain$) dataset. The results marked with $^\circ$ are cited from~\cite{Lee2022wildnet} and with $^*$ are cited from the respective paper. All results without any identifier are simulated. \textbf{Evaluation} is performed on the \textbf{SYNTHIA} and \textbf{GTA5 development sets} and on the \textbf{test$*$ data} of various real-world target datasets ($\tgt\!=\!\datateststar$). BM means benchmark. For our simulations we report mean values over three runs with different seeding. For models trained on SYNTHIA, we evaluate over 16 classes, as is common practice~\cite{Klingner2020c}. Best performance per group in bold face, second best underlined.}

\extrarowheight=\aboverulesep
    \addtolength{\extrarowheight}{\belowrulesep}
    \aboverulesep=0pt
    \belowrulesep=0pt
    \begin{tabular}{@{}cclccccccccccc@{}}
        \toprule[.9pt]
        &\multirow{1}{*}{\rot{\makecell[c]{\textbf{Enc.\hphantom{0}}}}} & \textbf{Method} &  \multirow{1}{*}{\makecell{$|\VECG{\theta}|$\\\textbf{($\cdot 10^6$)}} }&\multirow{1}{*}{\textbf{\makecell{ Single \\[-2pt] Source}}}&\multirow{1}{*}{\textbf{\makecell{ Frame \\[-2pt] Rate \\[-2pt] [fps]}}}&\multicolumn{8}{c}{\textbf{mIoU (\%) on}} \\
        \cmidrule{7-14}
       && & &  && $\csteststar$ & $\bddteststar$ & $\mvteststar$ & $\synthiadev$& $\gtavdev$  &$\acdcteststar$&$\kittiteststar$&\raisebox{-4.0pt}{\shortstack{\textbf{BM} \\\textbf{mean}}} \rule[-1.2ex]{0mm}{3.65ex} \\
        \midrule
        \parbox[t]{2mm}{\multirow{6}{*}{\rotatebox[origin=c]{90}{\small{{\textbf{Group 1}}}}}} &\parbox[t]{2mm}{\multirow{3}{*}{\rotatebox[origin=c]{90}{\footnotesize{\network{ResNet-50}}}}} &Baseline$^\bullet$ & 49.6&\ding{51}& 7.9 &28.36&25.16&27.24&-&-&-&-&-\\
        &&DRPC$^\bullet$~\cite{Yue2019}& 49.6 &\ding{55}&8.3& 35.65&31.53&32.74&-&-&-&-&-\\
        &&SAN+SAW*~\cite{Peng2022semanticaware}& 25.6 &\ding{51}&8.1& 38.92&\textbf{35.42}&34.52&-&29.16&-&-&-\\
        \cmidrule{2-14}
         &{\multirow{3}{*}{\rotatebox[origin=c]{90}{\footnotesize{\network{MiT-B2}}}}} &Baseline & 27.4 &\ding{51}&\textbf{12.0}& ${39.71}$ & ${29.76}$ & ${38.37}$ & $\mathbf{74.78}$ & ${37.83}$ & ${26.16}$ & $\underline{35.18}$ & ${44.09}$ \\

        &&Ours: \network{ReVT} \{\basemodel{4},\basemodel{5},\basemodel{6}\}   & 27.4 &\ding{51}&\textbf{12.0}&$\mathbf{41.09}$ & $\underline{35.18}$&$\underline{40.21}$&$\underline{71.59}$ & $\mathbf{40.88}$&$\mathbf{30.39}$ & ${34.64}$&$\mathbf{45.79}$\\

       &&Ours: \network{ReVT} \{\raisebox{.8pt}{\textcircled{\raisebox{-0.9pt}{1}}},\raisebox{.8pt}{\textcircled{\raisebox{-0.9pt}{4}}},\raisebox{.8pt}{\textcircled{\raisebox{-0.9pt}{6}}}\}   & 27.4 &\ding{51}&\textbf{12.0}&$\underline{40.91}$ & ${34.53}$ & $\mathbf{40.44}$ & ${71.45}$ & $\underline{39.87}$ & $\underline{30.13}$ & $\mathbf{35.29}$ & $\underline{45.44}$ \\
        \midrule[.9pt]
        \parbox[t]{2mm}{\multirow{7}{*}{\rotatebox[origin=c]{90}{\small{{\textbf{Group 2}}}}}} &\parbox[t]{2mm}{\multirow{4}{*}{\rotatebox[origin=c]{90}{\footnotesize{\network{ResNet-101}}}}} &Baseline$^\bullet$ & 68.6&\ding{51}& 7.9 &29.67&25.64&28.73&-&-&-&-&-\\
       & &DRPC$^\bullet$~\cite{Yue2019}& 68.6 &\ding{55}&5.3&37.58&34.34&34.12&-&-&-&-&-\\
        &&FSDR*~\cite{Huang2021}& 68.6 &\ding{55}&5.3&40.80&37.40&39.60&-&-&-&-&-\\
       & &SAN+SAW*~\cite{Peng2022semanticaware}& 44.6&\ding{51}& 5.3&40.87&35.98&37.26&-&30.79&-&-&-\\
     
        \cmidrule{2-14}
    
       &\parbox[t]{2mm}{\multirow{3}{*}{\rotatebox[origin=c]{90}{\footnotesize{\network{MiT-B3}}}}} &Baseline & 47.2&\ding{51}& \textbf{10.7}&${42.43}$ & ${33.33}$ & ${40.47}$ & $\mathbf{75.82}$ & ${41.53}$ & ${29.73}$ & ${35.91}$ & ${46.72}$ \\
           
         &&Ours: \network{ReVT} \{\basemodel{4},\basemodel{5},\basemodel{6}\}   & 47.2 &\ding{51}&\textbf{10.7}&$\mathbf{45.26}$ & $\mathbf{38.73}$ & $\underline{42.86}$ & ${73.12}$ & $\mathbf{44.99}$ & $\mathbf{35.27}$ & $\mathbf{36.42}$ & $\mathbf{48.99}$ \\
         &&Ours: \network{ReVT} \{\basemodel{1},\basemodel{4},\basemodel{6}\}   & 47.2 &\ding{51}&\textbf{10.7}&$\underline{44.97}$ & $\underline{38.65}$ & $\mathbf{43.00}$ & $\underline{73.16}$ & $\underline{44.42}$ & $\underline{35.16}$ & $\underline{36.31}$ & $\underline{48.84}$ \\
        \midrule[.9pt]
         \parbox[t]{2mm}{\multirow{3}{*}{\rotatebox[origin=c]{90}{\small{{\textbf{Group 3}}}}}} &\parbox[t]{2mm}{\multirow{3}{*}{\rotatebox[origin=c]{90}{\footnotesize{\network{MiT-B5}}}}} &Baseline & 84.7&\ding{51} &\textbf{9.7}& ${45.07}$ & ${35.19}$ & ${42.51}$ & $\mathbf{76.49}$ & $\underline{42.82}$ & ${30.81}$ & ${37.02}$ & ${48.42}$ \\
       
         & &Ours: \network{ReVT} \{\basemodel{4},\basemodel{5},\basemodel{6}\}   & 84.7 &\ding{51}&\textbf{9.7}&$\underline{45.08}$ & $\underline{39.62}$ & $\underline{43.99}$ & ${73.96}$ & $\mathbf{43.25}$ & $\underline{35.12}$ & $\underline{37.20}$ & $\underline{49.18}$ \\
         & &Ours: \network{ReVT} \{\raisebox{.8pt}{\textcircled{\raisebox{-0.9pt}{1}}},\raisebox{.8pt}{\textcircled{\raisebox{-0.9pt}{4}}},\raisebox{.8pt}{\textcircled{\raisebox{-0.9pt}{6}}}\}   & 84.7 &\ding{51}&\textbf{9.7} & $\mathbf{46.28}$ & $\mathbf{40.30}$ & $\mathbf{44.76}$ & $\underline{74.11}$ & ${42.74}$ & $\mathbf{35.75}$ & $\mathbf{37.86}$ & $\mathbf{49.64}$ \\
        
        \bottomrule[.9pt]
    \end{tabular}

  \label{tab:sota-synthia} 
  \vspace{-4.5mm}
\end{table*}

\subsection{SegFormer Block Diagrams}
\label{supp:block-diagramms}
In \textcolor{red}{Section}~\ref{sec:experiments} we investigated the effect of the re-parameterization on different network parts (cf.\ \autoref{tab:reparamlocation}) and block or layer types (cf.\ \autoref{tab:reparamlayertypes}). To give the reader a better idea of how the network is structured and where the individual block and layer types are located in the network, an hierarchically illustrated overview of the \network{SegFormer} architecture with an \network{MiTB5} encoder is given in \textcolor{red}{Figures}~\ref{fig:blockdiagramm-mitb5}, \ref{fig:blockdiagramm-olpe}, \ref{fig:blockdiagramm-trafoblock}, \ref{fig:blockdiagramm-mixffn}, and \ref{fig:blockdiagramm-attention}. 
\subsection{Discussion of Limitations}
Although modern methods for domain generalization provide good performance on completely unseen real data (after training on synthetic data), the performance still remains behind that of modern methods for unsupervised domain adaptation (UDA) [\textcolor{green}{47},\textcolor{green}{49}]. Such a comparison, however, is not entirely fair, since UDA methods employ unlabeled data from a target domain (typically Cityscapes) during the training process, which we intentionally avoid in domain generalization. Nevertheless, it should be noted that better performance on a specific target domain can be achieved, if samples from this domain are available during training. \par
Our proposed method cannot be applied advantageously to any already trained model, since the optimizer choice has a significant impact on the performance. To be fair, however, this is the case with all prior art methods as well. Most of them additionally extend the training process considerably, far beyond the choice of the optimizer~\cite{Choi2021,Yue2019,Huang2021,Peng2022semanticaware,Lee2022wildnet}. 

\subsection{Discussion of Ethical Implications}
Although well generalizing semantic segmentation has many civilian applications that provide great value to society, e.g., automated driving, robotics, and medical applications, this technology can also be used for military and surveillance applications. Research on better generalizing methods may also indirectly contribute to the improvement of these applications. \par
Another aspect to consider are biases in the employed datasets. Three of the five real datasets (Cityscapes, ACDC, KITTI) were captured in Central Europe, one in the USA (BDD100k), and only one contains data from all over the world (Mapillary Vistas). This may lead to biases regarding different ethnicities in the data, which were not investigated further in this paper. For the reported results on improved generalization from synthetic to real data, the biases may be negligible, but should be considered for possible real-world applications. \\[1em]
\begin{figure}[t]
\centering\includegraphics{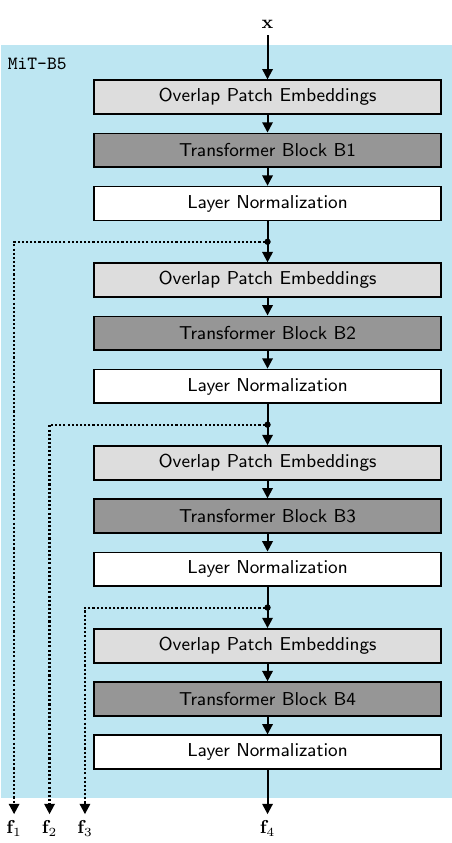}
  \caption{Overview of the \network{MiT-B5} encoder that is employed by the largest \network{SegFormer} model. This is the standard encoder employed in the segmenation model (cf. \autoref{fig:00_blockdiagramm_overview}).}
  \label{fig:blockdiagramm-mitb5}
\end{figure}
\begin{figure}[t]
  \centering
\includegraphics{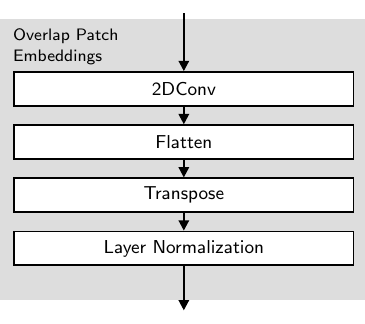}
  \caption{Overview of the overlap patch embeddings block that is employed in the \network{MiT} encoder (cf. \autoref{fig:blockdiagramm-mitb5} for \network{MiT-B5}).}
  \label{fig:blockdiagramm-olpe}
\end{figure}
\begin{figure}[t]
  \centering
  \includegraphics{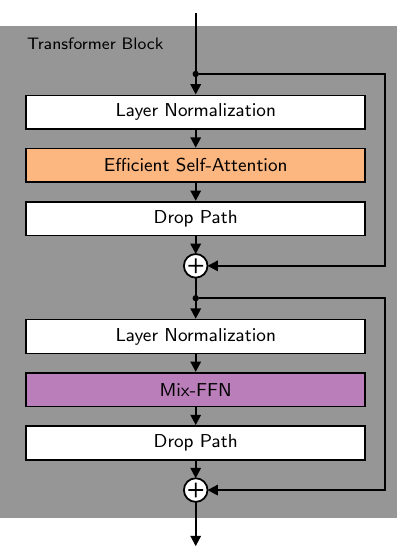}
  \caption{Overview of the transformer block that is employed in the \network{MiT} encoder (cf. \autoref{fig:blockdiagramm-mitb5} for \network{MiT-B5}).}
  \label{fig:blockdiagramm-trafoblock}
\end{figure}

\begin{figure}[t]
  \centering
  \includegraphics{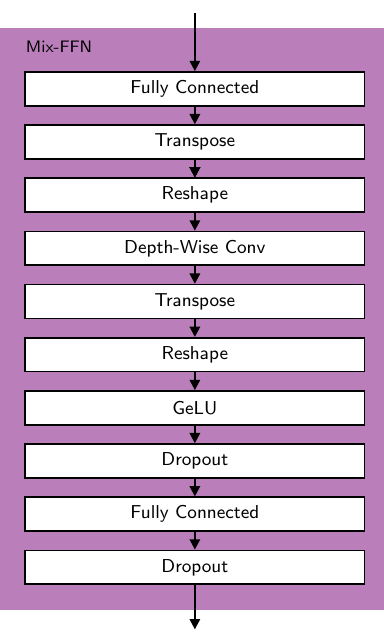}
  \caption{Overview of the Mix-FFN block that is employed in the transformer block (cf. \autoref{fig:blockdiagramm-trafoblock}).}
  \label{fig:blockdiagramm-mixffn}
\end{figure}
\begin{figure}[t]
  \centering
  \includegraphics{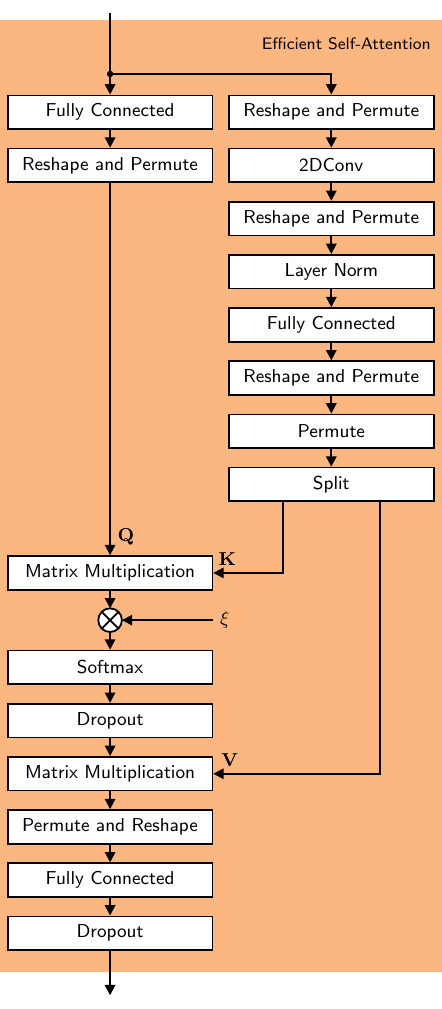}
  \caption{Overview of the efficient self-attention block that is employed in the transformer block (cf. \autoref{fig:blockdiagramm-trafoblock}). The fixed scaling factor $\xi$ is a hyperparameter and block-dependent.}
  \label{fig:blockdiagramm-attention}
\end{figure}
\newenvironment{literatur}{%
  \parskip2pt \parindent0pt 
  \def\lititem{\hangindent=0.7cm \hangafter1}}{%
  \par\ignorespaces}

\subsection*{Additional References}
\begin{literatur}
\lititem \small [47]\enspace Nikita Araslanov and Stefan Roth. Self-Supervised Augmentation Consistency for Adapting Semantic Segmentation. In \textit{Proc.\ of} CVPR, pages 15384–15394, virtual, June 2021.\par
\lititem \small [48]\enspace Kaiming He, Xiangyu Zhang, Shaoqing Ren, and Jian Sun.\ Deep Residual Learning for Image Recognition. In \textit{Proc.\ of} CVPR, pages 770–778, Las Vegas, NV, USA, June 2016 \par
\lititem \small [49]\enspace Lukas Hoyer, Dengxin Dai, and Luc Van Gool. DAFormer: Improving Network Architectures and Training Strategies for Domain-Adaptive Semantic Segmentation. In \textit{Proc.\ of} CVPR, pages 9924–9935, New Orleans, LA, USA, June 2022\par
\end{literatur}
\end{document}